\documentclass[Afour,sageh,times]{sagej}

\usepackage{moreverb,url}
\usepackage{amsmath} 
\usepackage{amssymb}  
\usepackage{subfigure}
\usepackage{multirow}
\usepackage{array,booktabs}
\usepackage{diagbox}
\usepackage{balance}
\usepackage{mathtools}
\usepackage{subfigure}
\usepackage{verbatim}
\usepackage{adjustbox}
\usepackage{algorithm}
\usepackage{algpseudocode}
\usepackage{romannum}
\usepackage{amssymb}
\usepackage{pifont}
\usepackage{hyperref}
\usepackage{enumitem}
\hypersetup{
  colorlinks=true, 
  urlcolor=blue,   
  linkcolor=blue,  
  citecolor=black  
}
\usepackage[table]{xcolor}
\usepackage{./rpm_packages/rpm_SIunits}
\usepackage{./rpm_packages/rpm_acronyms}
\usepackage{./rpm_packages/rpm_math}
\usepackage{./rpm_packages/rpm_misc}
\usepackage{soul,color}
\usepackage{multirow}
\usepackage{soul,color}
\usepackage{lipsum}
\usepackage{tikz} 
\usepackage{MnSymbol}

\newcommand\BibTeX{{\rmfamily B\kern-.05em \textsc{i\kern-.025em b}\kern-.08em
T\kern-.1667em\lower.7ex\hbox{E}\kern-.125emX}}

\def\volumeyear{2024}

\usepackage{bbding,pifont}

\begin{document}

\runninghead{Jung \emph{et al.}}

\title{HeLiPR: Heterogeneous LiDAR Dataset for inter-LiDAR Place Recognition under Spatiotemporal Variations}

\author{Minwoo Jung\affilnum{1}, Wooseong Yang\affilnum{1}, Dongjae Lee\affilnum{1}, Hyeonjae Gil\affilnum{1}, Giseop Kim\affilnum{2}, Ayoung Kim\affilnum{1}}

\affiliation{\affilnum{1}Dept. of Mechanical Engineering, SNU, Seoul, S. Korea \\
\affilnum{2}NAVER LABS, Gyeonggi-do, S. Korea}

\corrauth{Ayoung Kim, Dept. of Mechanical Engineering, SNU, Seoul, S. Korea}
\email{ayoungk@snu.ac.kr}

\begin{abstract}
Place recognition is crucial for robot localization and loop closure in \ac{SLAM}. \ac{LiDAR}, known for its robust sensing capabilities and measurement consistency even in varying illumination conditions, has become pivotal in various fields, surpassing traditional imaging sensors in certain applications. Among various types of LiDAR, spinning LiDARs are widely used, while non-repetitive scanning patterns have recently been utilized in robotics applications. Some LiDARs provide additional measurements such as reflectivity, \ac{NIR}, and velocity from \ac{FMCW} LiDARs. Despite these advances, there is a lack of comprehensive datasets reflecting the broad spectrum of LiDAR configurations for place recognition. To tackle this issue, our paper proposes the HeLiPR dataset, curated especially for place recognition with heterogeneous LiDARs, embodying spatiotemporal variations. To the best of our knowledge, the HeLiPR dataset is the first heterogeneous LiDAR dataset supporting inter-LiDAR place recognition with both non-repetitive and spinning LiDARs, accommodating different \ac{FOV}s and varying numbers of rays. The dataset covers diverse environments, from urban cityscapes to high-dynamic freeways, over a month, enhancing adaptability and robustness across scenarios. Notably, HeLiPR includes trajectories parallel to MulRan sequences, making it valuable for research in heterogeneous LiDAR place recognition and long-term studies. The dataset is accessible at \url{https://sites.google.com/view/heliprdataset}.

\end{abstract}


\maketitle

\section{Introduction}
Place recognition is an essential task in robotics, involving the ability to identify whether a place has been visited before or not. The significance of this task stems from its role as an initial step towards localization and its contribution to enabling loop closure in SLAM. Traditionally, it has been accomplished by searching a query image within a database using image sensors \citep{zhang2010understanding, arandjelovic2016netvlad, 9635907}.
However, recent advancements have facilitated the adoption of \ac{LiDAR} for place recognition, attributable to its enhanced geometric sensing capabilities. \ac{LiDAR}-based place recognition has been gaining attraction thanks to its capacity to measure the range precisely, and distinct from image sensors, \ac{LiDAR} has the advantage of capturing geometric structures with illumination invariance. Conventionally, \ac{LiDAR} descriptors \citep{kim2021scan, 9981308, 9462410} are generated from the scan and subsequently used to ascertain the presence or absence of a place through comparison with a comprehensive set of descriptors. While place recognition can be replaced using \ac{GPS}, it has limitations in environments where signals are weak. LiDAR overcomes these challenges with high-resolution spatial data, enabling accurate place recognition even in GPS-denied areas, underscoring its importance in complex navigation tasks.

\begin{figure*}[!t]
    \centering
    \includegraphics[width=.98\textwidth]{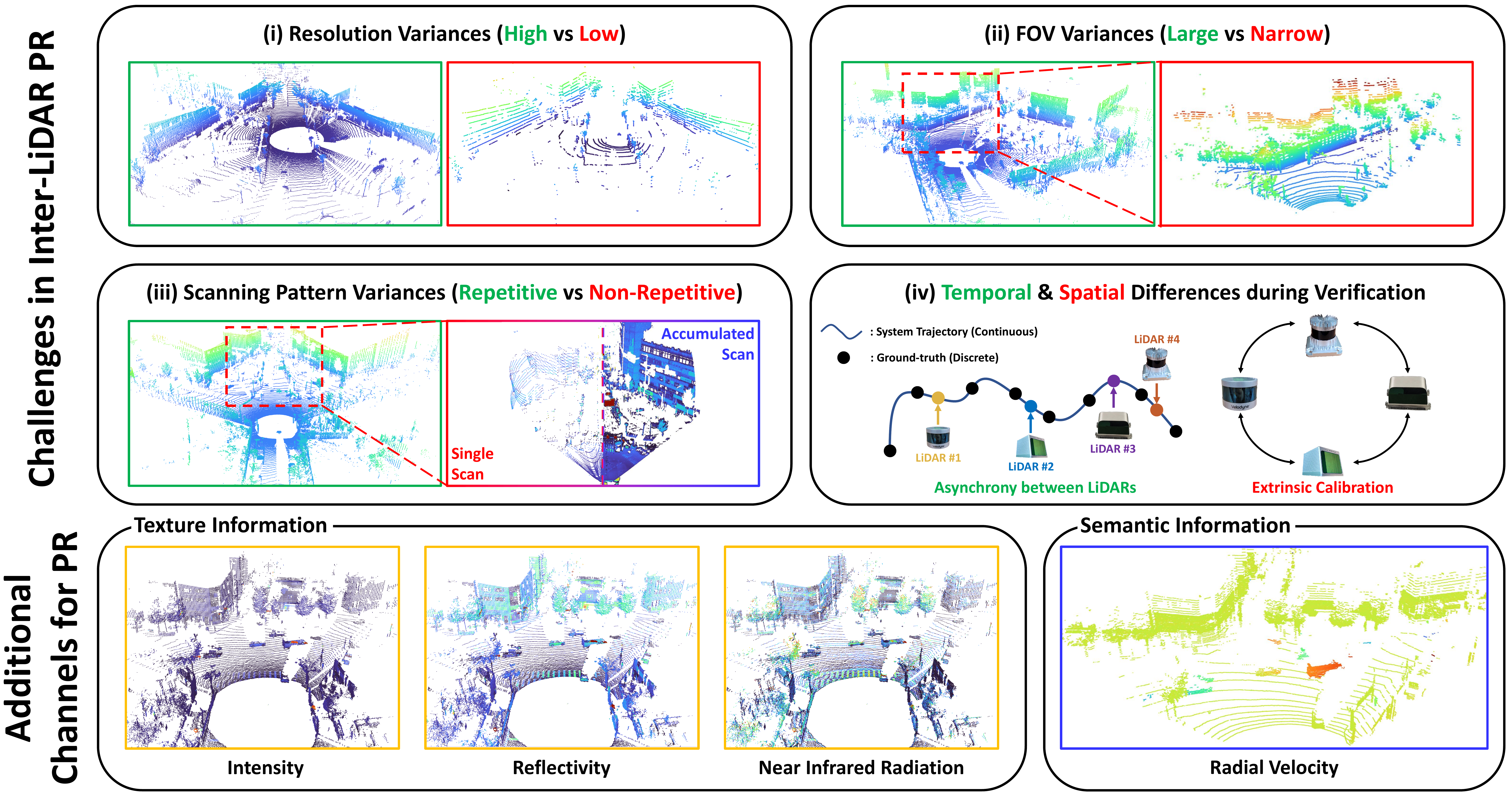}
    \vspace{-2mm}
    \captionof{figure}{(Top row) 
    LiDAR place recognition challenges (i) Variance in resolution between high and low ray count LiDARs affects sensing abilities. (ii) While some LiDARs perform 360-degree scans, others have limited FOV due to occlusion or sensor limitations. (iii) Most LiDARs scan the repetitive area, whereas non-repetitive LiDAR densely scan by stacking individual scans. However, each scan tends to be sparse, as depicted in the left red box. (iv) Ground truth, crucial for executing LiDAR place recognition, is challenging to determine due to varying LiDAR coordinates and scan acquisition times. (Bottom row) HeLiPR dataset provides heterogeneous LiDARs and additional channels, thereby granting opportunities to utilize texture information from LiDAR.}
    \label{fig:overview}
    \vspace{-5mm}
\end{figure*}

With the advancement of place recognition, the hardware capabilities of \ac{LiDAR} have also evolved significantly. For instance, specific \ac{LiDAR}s deploy non-repetitive scanning patterns to achieve dense mapping, thus deviating from traditional spinning \ac{LiDAR}s. Additionally, some \ac{LiDAR}s feature a more significant number of rays, surpassing the conventional 16 or 32-ray configurations, and incorporate additional channels such as reflectivity and \ac{NIR}. More recently, the advent of FMCW \ac{LiDAR} has made it possible to measure relative velocity along the radial direction utilizing the Doppler effect, commonly called velocity measurement. Considering these developments in LiDARs, place recognition with non-repetitive scanning pattern \ac{LiDAR}s \citep{2023STD, 10388464} has also been pursued. Furthermore, studies \citep{wang2020intensity, shan2021robust, chen2020rss} that leverage the information offered by the additional channels in \ac{LiDAR} have also emerged.

Nevertheless, despite these advancements, there currently exists a scarcity of datasets incorporating diverse combinations of \ac{LiDAR}s for place recognition. This shortfall highlights a gap in the availability of benchmark datasets for validating place recognition operating with heterogeneous \ac{LiDAR}s. Several datasets \citep{9197298, wild, Geiger2012CVPR} are conducive for tasks involving place recognition, although they are equipped solely with a spinning \ac{LiDAR}. On the other hand, while there are datasets inclusive of multiple \ac{LiDAR}s, these predominantly feature spinning \ac{LiDAR}s \citep{jjeong-2019-ijrr, RadarRobotCarDatasetICRA2020, agarwal2020ford, hsu2021urbannav}, or they comprise heterogeneous \ac{LiDAR}s that are ill-suited for place recognition \citep{qingqing2022multi, helmberger2022hilti, jung2023asynchronous}.

This paper introduces the HeLiPR dataset, a unique heterogeneous LiDAR dataset for place recognition, encapsulating spatiotemporal variations. Regarding environmental diversity, our dataset was acquired over 37 days, with data collection occurring 3 to 4 times. This acquisition provides a variety of environments encompassing a narrow residential area, urban cityscape, and environments with high dynamic change.
One of the key aspects of our dataset is the focus on inter-LiDAR place recognition, which refers to the challenge of using multiple and diverse LiDARs both within individual sessions, known as intra-session, and across different sessions, referred to as inter-session.
The complex challenges associated with this form of place recognition are comprehensively represented in \figref{fig:overview}. The HeLiPR dataset thoroughly encompasses each highlighted challenge, demonstrating its applicability in the field. In addition, the HeLiPR dataset includes trajectories similar to sequences acquired from MulRan \citep{9197298}, enabling another heterogeneous LiDAR and long-term place recognition with a term of four years. The salient contributions of the HeLiPR dataset are as follows:


\hypersetup{
    colorlinks=true,
    citecolor=red
}

\begin{table*}[]
\centering
\caption{Dataset comparison for LiDAR-based place recognition. The number of channels refers to the additional channels, excluding intensity. When the dataset contains non-identical paths that partially overlap with other sessions, it has inter-session difference. Furthermore, the $\filledstar$ represents the spatial scale based on sequence length.}
\label{tab:dataset}
\resizebox{\textwidth}{!}{%
\begin{tabular}{c|ccc|cccc|cc} \toprule
\multirow{2}{*}{Name} &
  \multicolumn{3}{c|}{LiDAR} &
  \multicolumn{4}{c|}{Loop Closure} &
  \multirow{2}{*}{Spatial Scale}&
  \multirow{2}{*}{Total Distance} \\ 
 &
  \# Spinning &
  \# Solid State &
  \# Channels &
  Intra-session &
  Inter-session &
  Temporal Diversity&
  Inter-session Differnce &
   &
   \\ \midrule
\begin{tabular}[c]{@{}c@{}}KITTI\\ \cite{Geiger2012CVPR} \end{tabular}                      &  1&  \ding{55}&  \ding{55}&  \ding{51}&  \ding{55}&  \ding{55}&  \ding{55}&  $\filledstar$ $\filledstar$ & 44 km \\ 
\begin{tabular}[c]{@{}c@{}}Complex Urban\\ \cite{jjeong-2019-ijrr} \end{tabular}            &  2&  \ding{55}&  \ding{55}&  \ding{51}&  \ding{51}&  15 months&  \ding{51} & $\filledstar$ $\filledstar$ $\filledstar$ & 190 km \\
\begin{tabular}[c]{@{}c@{}}Oxford Rdar\\ \cite{RadarRobotCarDatasetICRA2020} \end{tabular}  &  2&  \ding{55}&  \ding{55}&  \ding{51}&  \ding{51}&  \ding{55}&  \ding{55}&  $\filledstar$ $\filledstar$ $\filledstar$& 280 km\\
\begin{tabular}[c]{@{}c@{}}Ford Multi AV\\ \cite{agarwal2020ford} \end{tabular}             &  4&  \ding{55}&  \ding{55}&  \ding{51}&  \ding{51}&  4 months&  \ding{55}& $\filledstar$ $\filledstar$ $\filledstar$ &  198 km \\
\begin{tabular}[c]{@{}c@{}}MulRan\\ \cite{9197298} \end{tabular}                            &  1&  \ding{55}&  \ding{55}&  \ding{51}&  \ding{51}&  2 months & \ding{51} & $\filledstar$ $\filledstar$ $\filledstar$ & 123 km \\
\begin{tabular}[c]{@{}c@{}}NTU VIRAL\\ \cite{Viral} \end{tabular}                           &  2&  \ding{55}&  2 &  \ding{55}&  \ding{55}&  \ding{55}&  \ding{55}& $\filledstar$  & 1.9 km \\
\begin{tabular}[c]{@{}c@{}}Boreas\\ \cite{Boreas} \end{tabular}                             &  1&  \ding{55}&  \ding{55}&  \ding{51}&  \ding{51}&  12 months&  \ding{55}& $\filledstar$ $\filledstar$ $\filledstar$ &  350 km \\
\begin{tabular}[c]{@{}c@{}}Pohang Canal\\ \cite{Pohang} \end{tabular}                       &  3&  \ding{55}&  2 &  \ding{55}&  \ding{51}&  1 month&  \ding{55}& $\filledstar$ $\filledstar$ & 45 km  \\
\begin{tabular}[c]{@{}c@{}}Wild Place\\ \cite{wild} \end{tabular}                           &  1&  \ding{55}&  \ding{55}&  \ding{51}&  \ding{51}&  14 months&  \ding{51} & $\filledstar$ $\filledstar$ & 33 km \\
\begin{tabular}[c]{@{}c@{}}Hilti 2021\\ \cite{helmberger2022hilti} \end{tabular}            &  1&  1&  2&  \ding{55}&  \ding{55}&  \ding{55}&  \ding{55}& $\filledstar$ &  2.1 km \\
\begin{tabular}[c]{@{}c@{}}Tiers\\ \cite{qingqing2022multi} \end{tabular}                   &  3&  3&  2 &  \ding{55}&  \ding{55}&  \ding{55}&  \ding{55}& $\filledstar$ &  2 km \\ \midrule
\textbf{HeLiPR}                                                                             &  \textbf{2}&  \textbf{2}&  \textbf{3} &  \ding{51}&  \ding{51}&  \textbf{1 + 53 months} & \ding{51} & $\filledstar$ $\filledstar$ $\filledstar$ & \textbf{164 km} \\ \bottomrule
\end{tabular}%
}
\vspace{-2mm}
\end{table*}

\hypersetup{
    colorlinks=true,
    citecolor=black
}

\begin{enumerate}
    \item The HeLiPR dataset includes heterogeneous LiDARs, with OS2-128, VLP-16, Livox Avia, and Aeries II, while most of the existing dataset involves only spinning LiDARs. This configuration can underscore the impact of disparities in resolution and scanning patterns. Their additional channels, such as \ac{NIR}, reflectivity, and radial velocity, pave the way for novel strategies in place recognition.
    \item The HeLiPR dataset tackles heterogeneous LiDAR place recognition. Based on our benchmark results, the HeLiPR dataset underscores the growing need for dedicated research in heterogeneous inter-LiDAR place recognition. Furthermore, this dataset plays a significant role in facilitating and guiding essential research explorations in this field.
    \item The HeLiPR dataset captures diverse environments monthly, from residential to dynamic urban areas. Moreover, trajectories akin to those in MulRan enable heterogeneous LiDAR place recognition and support long-term research spanning four years. This broad spectrum of data acquisition positions HeLiPR as a pivotal tool for generalizing place recognition across varied scenarios.
    \item The HeLiPR dataset provides individual LiDAR ground truth corresponding to the acquisition time of each LiDAR. This accurate ground truth, which also considers spatial relationships, facilitates more accessible validation and improves the reliability of place recognition.
\end{enumerate}
\section{Related works}
This section presents an overview of LiDAR datasets pertinent to our research. A summary is provided in \tabref{tab:dataset}.

The KITTI dataset \citep{Geiger2012CVPR}, gathered using a carlike vehicle, represents a mid-sized cityscape. While it facilitates intra-session place recognition, the dataset falls short in supporting inter-session place recognition, with data acquisition solely reliant on a single HDL-64E. On the other hand, the Oxford Robotcar Radar Dataset \citep{RadarRobotCarDatasetICRA2020}, which shares a similar environment with KITTI, introduces the possibility for inter-session place recognition. However, even though multiple LiDARs are incorporated, all are of the spinning type. The Ford Multi-AV Dataset \citep{agarwal2020ford} stands out due to its extensive trajectory covering a range of environments from urban to vegetated, including tunnels, and showcasing seasonal changes. Similarly, Boreas \citep{Boreas} meets the conditions necessary for intra and inter-session place recognition. However, each sequence from Boreas and the Ford Multi-AV dataset consists of similar paths, reducing the complexity in inter-session place recognition. The Complex Urban Dataset \citep{jjeong-2019-ijrr} and UrbanNav Dataset \citep{hsu2021urbannav}, both situated within urban environments, lean more towards intra-session place recognition, offering limited avenues for inter-session recognition. The Wild Places \citep{wild} stands apart by ensuring both intra-session and inter-session place recognition, factoring in temporal variations. Nevertheless, it focuses on unstructured terrains and employs a single spinning LiDAR. Unlike the previous dataset, the Pohang Canal dataset \citep{Pohang} utilizes multiple LiDARs. However, sessions for inter-session place recognition are not adequate as the trajectory of all sessions has an identical path. The NTU VIRAL dataset also exploits multiple LiDARs; however, its primary focus on \ac{UAV} localization, particularly in smaller areas for maintaining the tracking of the Leica laser system, tends to overshadow its application in place recognition.

\begin{table}[!t]
\caption{The heterogeneous LiDARs utilized in Our Dataset}
\label{tab:lidar_config}
\resizebox{\columnwidth}{!}{%
\begin{tabular}{c|c|c|c|c|c} \toprule
Sensor      & Manufacture & Model     & Channel & FOV (H$\times$ V)  & Range \\ \midrule
Spinning    & Ouster      & OS2-128   & 128     & $360^\circ\times22.5^\circ$ & 200 m  \\
Spinning    & Velodyne    & VLP-16    & 16      & $360^\circ\times30^\circ$  & 100 m  \\
Solid state & Livox       & Avia      & 6       & $70^\circ\times77^\circ$    & 450 m  \\
Solid state & Aeva        & Aeries II & 64      & $120^\circ\times19.2^\circ$ & 150 m \\ \bottomrule
\end{tabular}%
}
\vspace{-5mm}
\end{table}

Many of the datasets mentioned above primarily rely on spinning LiDARs. More recent datasets, such as Tiers \citep{qingqing2022multi}, Hilti 2021 \citep{helmberger2022hilti}, and City \citep{jung2023asynchronous} dataset, have begun to incorporate heterogeneous LiDARs. The Tiers and Hilti 2021 datasets feature short-term indoor and outdoor data collection using carlike vehicles and handheld systems. Similarly, the City dataset captures urban areas using a vehicle-based system. Although these datasets employ heterogeneous LiDARs, their primary focus is on \ac{SLAM}. As a result, they tend to have relatively short paths, which means that revisits are either minimal or non-existent in sequences, rendering intra-session place recognition unfeasible. Additionally, the lack of overlap in their sequences means these datasets are unsuitable for validating inter-session place recognition.

The HeLiPR dataset distinguishes itself from others by showcasing diverse LiDARs, encompassing the OS2-128, VLP-16, Livox Avia, and Aeva Aeries II, each with unique attributes. These sensors capture data channels such as NIR, reflectivity, and radial velocity, ushering in new avenues for inventive place recognition. Significantly, the HeLiPR dataset captures each sequence over a month, supplying rich environments that support both intra-session and inter-session place recognition. Furthermore, the HeLiPR dataset trajectories resonate with sequences derived from MulRan, thus promoting research in heterogeneous LiDAR place recognition and offering an extended temporal perspective. Conclusively, every sequence in HeLiPR illustrates a vast environment with variations.
\section{System Overview}

\begin{figure}[!t]
    \centering
    \includegraphics[width=\columnwidth]{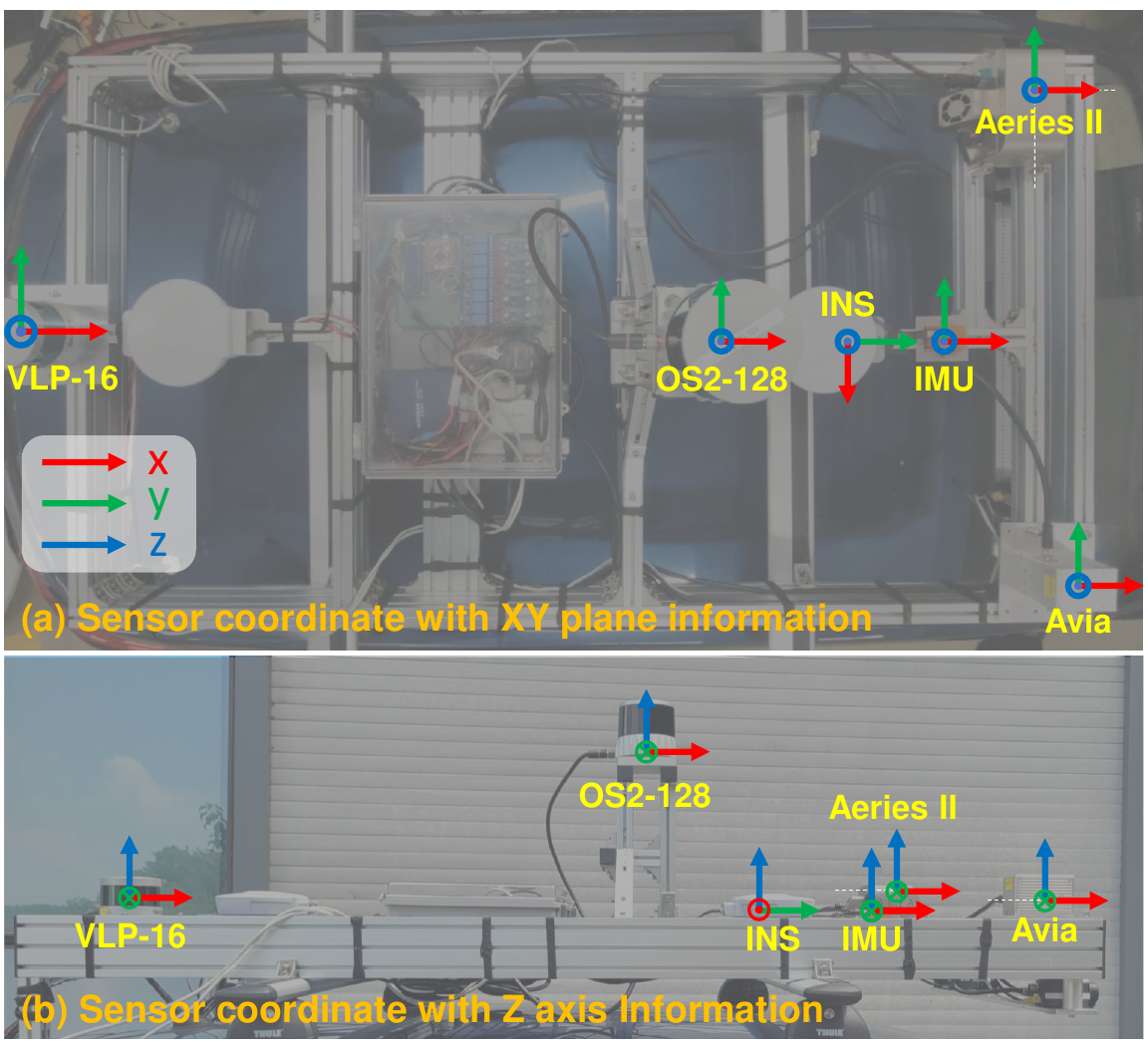}
    \caption{Sensors coordinate information between sensors. (a) and (b) represent the transformation with the xy-plane and z-axis. After the extrinsic calibration, the inter-sensor transformation can be found in the \texttt{Calibration} folder.}
    \label{fig:SensorConfig}
    \vspace{-5mm}
\end{figure}

\subsection{System Configuration}
Our system comprises four distinct LiDARs, as depicted in \tabref{tab:lidar_config} and \figref{fig:SensorConfig}. The spinning OS2-128 LiDAR is mounted at the center of the system, elevated to allow dense scanning without any occlusion. 
In contrast, the spinning LiDAR, VLP-16, experiences particular self-occlusion because of its proximity to the front box and surrounding sensors. This self-occlusion leads to an inability to fully scan the front view, impacting the data collection and operational limitations. Also, due to inherent hardware constraints, it casts a significantly smaller ray than the OS2-128, enabling only a peripheral scan of its surroundings.
The remaining LiDARs, Livox Avia and Aeva Aeries II are oriented to scan the front view of the vehicle, and each presents unique limitations. In the case of the Avia, its unconventional scanning patterns deviate from traditional spinning LiDARs; thus, direct comparison with them is challenging. However, Avia can construct dense maps with accumulating non-repetitive scans based on the relative transformation between scans.
The Aeries II also presents a narrow horizontal FOV. 
Even if this LiDAR has the advantage of detecting radial velocities of points, \ac{FMCW} technology introduces noise into the range measurements. Among several range configurations, we choose the configuration with maximum range of 150 m. This choice is made since a longer-range configuration leads to more noise, which could compromise accuracy.
This combination of LiDARs, with their unique scanning patterns, allows for an intriguing exploration in place recognition, including dealing with occlusion scenarios and contrasting low versus high resolution as shown in \figref{fig:overview}. Furthermore, leveraging the characteristics of these LiDARs could significantly enhance place recognition in environments characterized by substantial dynamics or rich textures. The additional channels offered by each sensor can be found in \figref{fig:FileSystem}, and we have identically configured all the LiDAR sensors to operate at a frequency of 10 Hz.

In addition to the LiDARs, our system incorporates two types of inertial sensors, the \ac{IMU} and the \ac{INS}. These devices provide a means to determine the temporal and spatial relationships within the asynchronous LiDAR system. We employ the Xsens MTi-300, which measures inertial information at 100 Hz. We use the SPAN-CPT7 coupled with a dual VEXXIS GNSS-501 antenna to establish a baseline for the vehicle system. All baselines are achieved at a frequency of 50 Hz using RTK GPS and \ac{INS}. Due to each sensor acquiring measurements in its own coordinate system, an extrinsic calibration process is necessary to integrate all the data into a standard coordinate system. This ensures consistency and accuracy across various measurements.

\begin{figure}[!t]
  \centering
    \begin{minipage}{0.95\columnwidth} 
    \centering
    \subfigure[Calibration Trajectory]
    {
    \includegraphics[height=0.5\columnwidth]{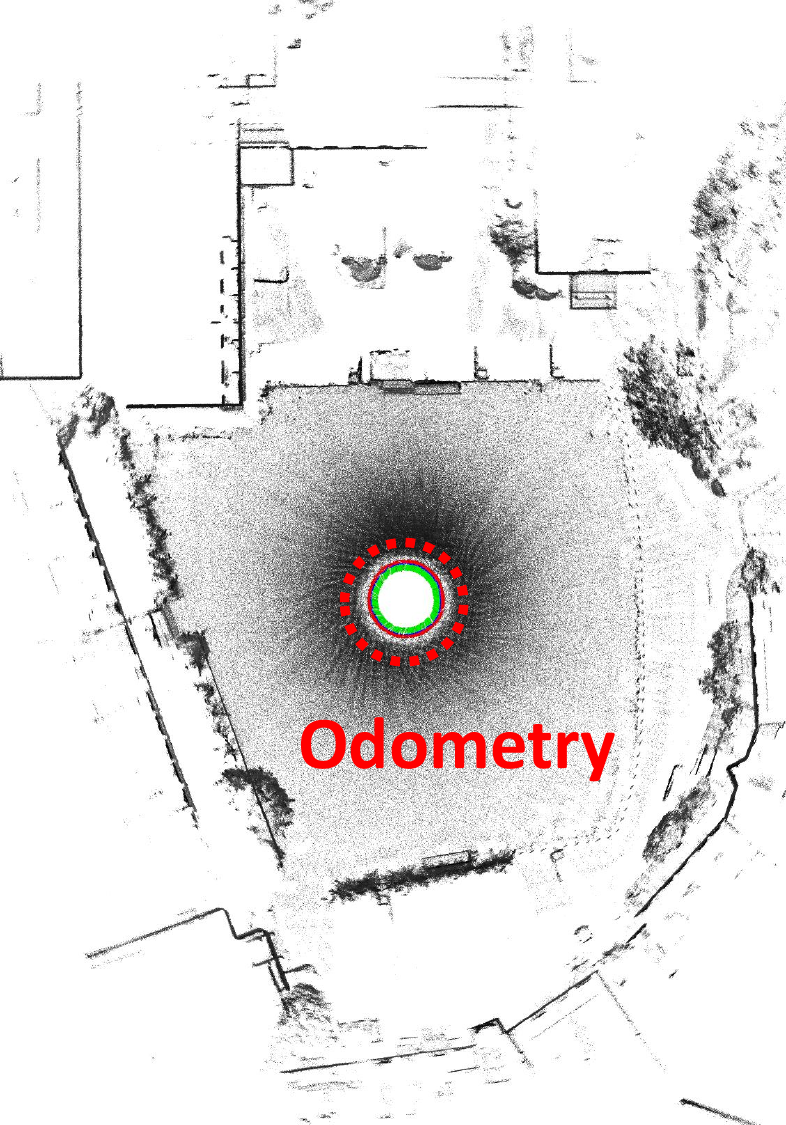}
    \label{fig:LiDARCalib_a}
    }
    \subfigure[Entire Scans from LiDARs]
    {
    \includegraphics[height=0.49\columnwidth]{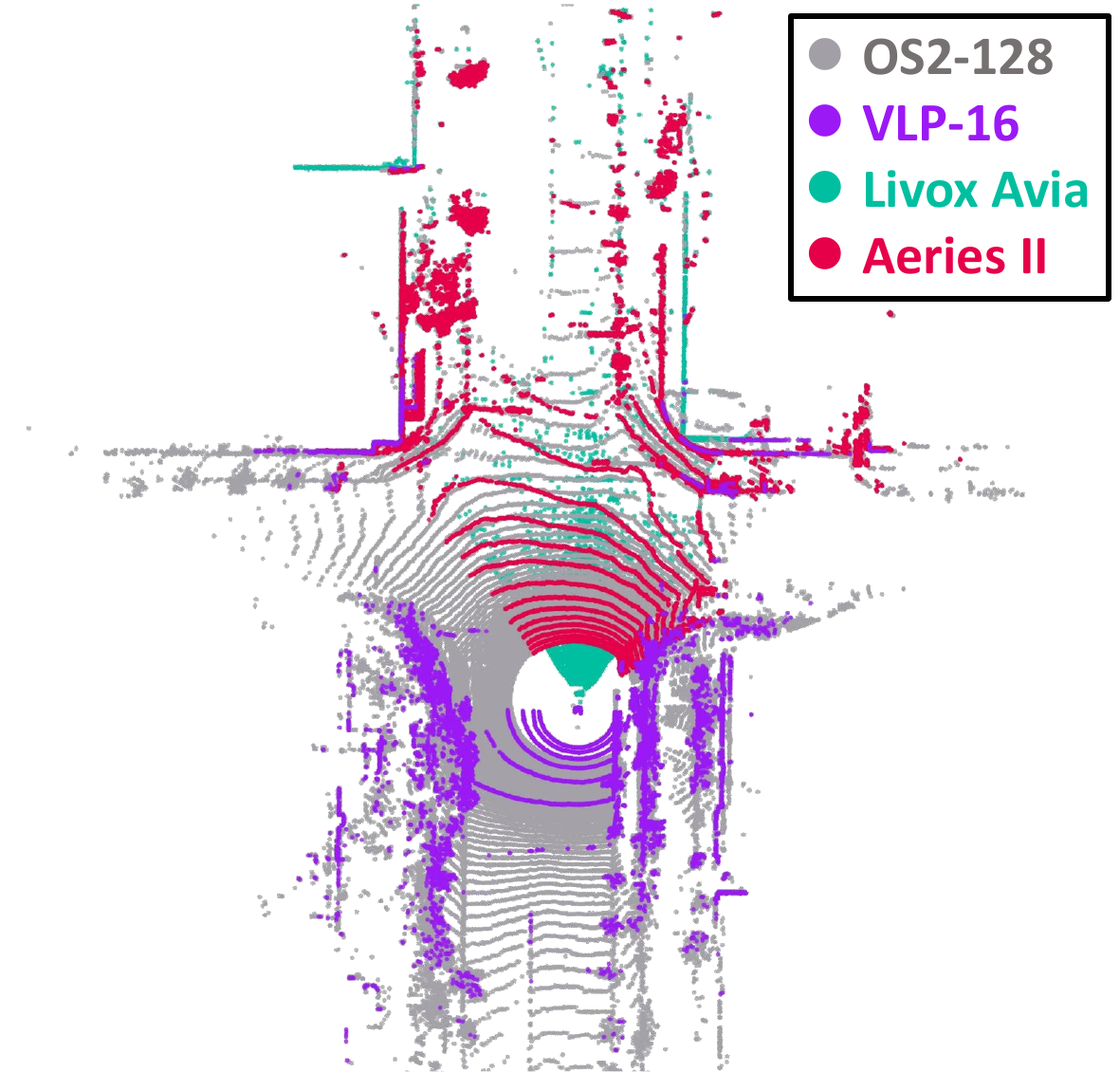}
    \label{fig:LiDARCalib_b}
    }
    \end{minipage}
  \vspace{-4mm}
  \caption{\subref{fig:LiDARCalib_a} Extrinsic calibration trajectory: A circular path used for map construction and calibration purposes. \subref{fig:LiDARCalib_b} Post-calibration LiDAR alignment: A sky plot view illustrating the contours overlap between individual LiDAR scans.}
  \label{fig:overall_system}
  \vspace{-6mm}
\end{figure}

\subsection{Sensor Calibration}
\label{sec:sensorCalib}
For simplicity, we employ symbols to represent the coordinate systems: $L$ corresponds to the LiDAR, $N$ signifies the INS, $I$ is used for the IMU, and $W$ indicates the world system. 
\subsubsection{Multiple LiDAR Extrinsic Calibration}
To calculate the extrinsic calibration between LiDARs, we utilize the existing calibration method \citep{9779777}. In this method, the trajectories from each LiDAR are obtained and updated through batch optimization with scans from each LiDAR. After that, based on the updated trajectories and the initial extrinsic calibration parameters, batch optimization with multiple LiDARs is re-performed to calculate an accurate extrinsic calibration.
To implement this method properly, a specified procedure is followed. The initial step involves moving the system a minimum distance to ensure trajectory accuracy. A complete 360-degree rotation follows this to facilitate the capture of loop closure for narrow \ac{FOV} LiDARs. In the case of vehicles, given their inability to rotate in place, the system proceeded with a circular trajectory as shown in \figref{fig:LiDARCalib_a}.
Furthermore, considering the potential distortion of LiDAR that might occur during motion, we stop the movement for 10 seconds during a motion and acquire a total of 30 scans with stationary. Lastly, considering the sparsity of a single scan from the Livox Avia, LiDAR scans are accumulated in a stationary condition. The initial extrinsic calibration is established using the CAD model. Additionally, the odometry for each LiDAR is obtained using Direct LiDAR Odometry \citep{chen2022direct}. As depicted in Figure \ref{fig:LiDARCalib_b}, it is clear that all the LiDARs are accurately aligned due to the precise extrinsic calibration.
\subsubsection{IMU-LiDAR Extrinsic Calibration}
The extrinsic calibration between the IMU and the LiDAR commences with the CAD model serving as the initial estimation. The extrinsic calibration of the IMU-LiDAR ($\textbf{T}^L_I$) is subsequently computed using LiDAR-Inertial Odometry \citep{9697912}. It is updated while transforming the LiDAR scan to IMU coordinates and calculating the point-to-plane distance relative to the global map.
However, achieving 6-DOF motion with a vehicle can prove challenging, which may adversely impact the accuracy of the extrinsic calibration between them. To mitigate this issue, the extrinsic calibration is updated with a minimal covariance, ensuring no significant deviation from the initial estimation.
The entire process is executed based on the \texttt{Roundabout01} sequence, leading to the calculation of extrinsic calibration between the OS2-128 and the IMU. The reason for selecting these two sensors is that they are positioned colinearly, resulting in an almost zero distance between one axis. Additionally, as the two sensors share the same axis, the initial estimate of these parameters remains the most accurate among all the LiDARs.

\begin{table*}[!t]
\caption{The Description for Each Sequence.}
\label{tab:data_description}
\resizebox{\textwidth}{!}{%
\begin{tabular}{c|c|cccccccccccc} \toprule
\multirow{3}{*}{Sequence Name} & \multirow{3}{*}{Characteristics}             & \multicolumn{12}{c}{Sequence Index}                                                                          \\
                               &                                         & \multicolumn{3}{c}{01}                   & \multicolumn{3}{c}{02}                   & \multicolumn{3}{c}{03}   
                                 & \multicolumn{3}{c}{04} 
                               \\
                               &                                         & Date & Duration & Distance                & Date & Duration & Distance                & Date & Duration & Distance &
                               Date & Duration & Distance\\ \midrule
\texttt{Roundabout}                     & Various rotation variations           & 2023-07-16     &  2730 \unit{s}      & \multicolumn{1}{c|}{9040 \unit{m}} & 2023-08-01     & 2085 \unit{s}          & \multicolumn{1}{c|}{7447 \unit{m}} &   2023-08-13   &  2515 \unit{s}      & \multicolumn{1}{c|}{9262 \unit{m}} & -     & -   & \multicolumn{1}{c}{-}       \\
\texttt{Town}                           & FOV issue in narrow areas               & 2023-07-18     &  2414 \unit{s}        & \multicolumn{1}{c|}{7832 \unit{m}} & 2023-07-31     & 2689 \unit{s}        & \multicolumn{1}{c|}{8203 \unit{m}} & 2023-08-14     & 2528 \unit{s}   & \multicolumn{1}{c|}{8903 \unit{m}} & -     & -   & \multicolumn{1}{c}{-}        \\
\texttt{Bridge}                         & Similar scenes and dynamic objects & 2023-07-17     &  2144 \unit{s}        & \multicolumn{1}{c|}{23056 \unit{m}} &  2023-07-31    & 2562 \unit{s}         & \multicolumn{1}{c|}{14615 \unit{m}} & 2023-08-14      &  2009 \unit{s}      & \multicolumn{1}{c|}{19400 \unit{m}} & 2023-08-21      &  3033 \unit{s}      & \multicolumn{1}{c}{22958  \unit{m}}       \\ \bottomrule
\end{tabular}%
}
\vspace{-3mm}
\end{table*}

\subsubsection{INS-IMU Extrinsic Calibration}
The extrinsic calibration between the INS and IMU is conducted using MA-LIO \citep{jung2023asynchronous}. This method is particularly effective for asynchronous LiDARs.
As the trajectory from MA-LIO is aligned to the IMU coordinate system, the subsequent hand-eye calibration between the INS and IMU can be executed. Specifically, the relative transformation, or $\Delta\textbf{T}^N$, between $\textbf{T}^N$ at two distinct timestamps $t^N_i$ and $t^N_j$ is determined. Similarly, the relative transformation, $\Delta\textbf{T}^I$, between $\textbf{T}^I$ at timestamps $t^I_i$ and $t^I_j$ is ascertained. Considering the non-coincident time, $t^N$, and $t^I$, we synchronize the acquisition time across both sensors to minimize time discrepancies. 
Then, the extrinsic calibration between the INS and IMU is achieved via the equation $\Delta\textbf{T}^N \textbf{T}_N^I = \textbf{T}_N^I \Delta\textbf{T}^I$. 
We employ $15000$ transformations from the \texttt{Roundabout01} sequence to carry out the calibration, with the maximum time difference registering at approximately 1 \unit{ms}. Moreover, it is worth noting that the fidelity of hand-eye calibration is inherently reliant on the precision of the MA-LIO. As such, we employ the CAD model specifically for the z-axis translation, which is most susceptible to errors in LIO. For all other components, we use the results from the hand-eye calibration.

\begin{figure}[!t]
    \centering
    \includegraphics[width=\columnwidth]{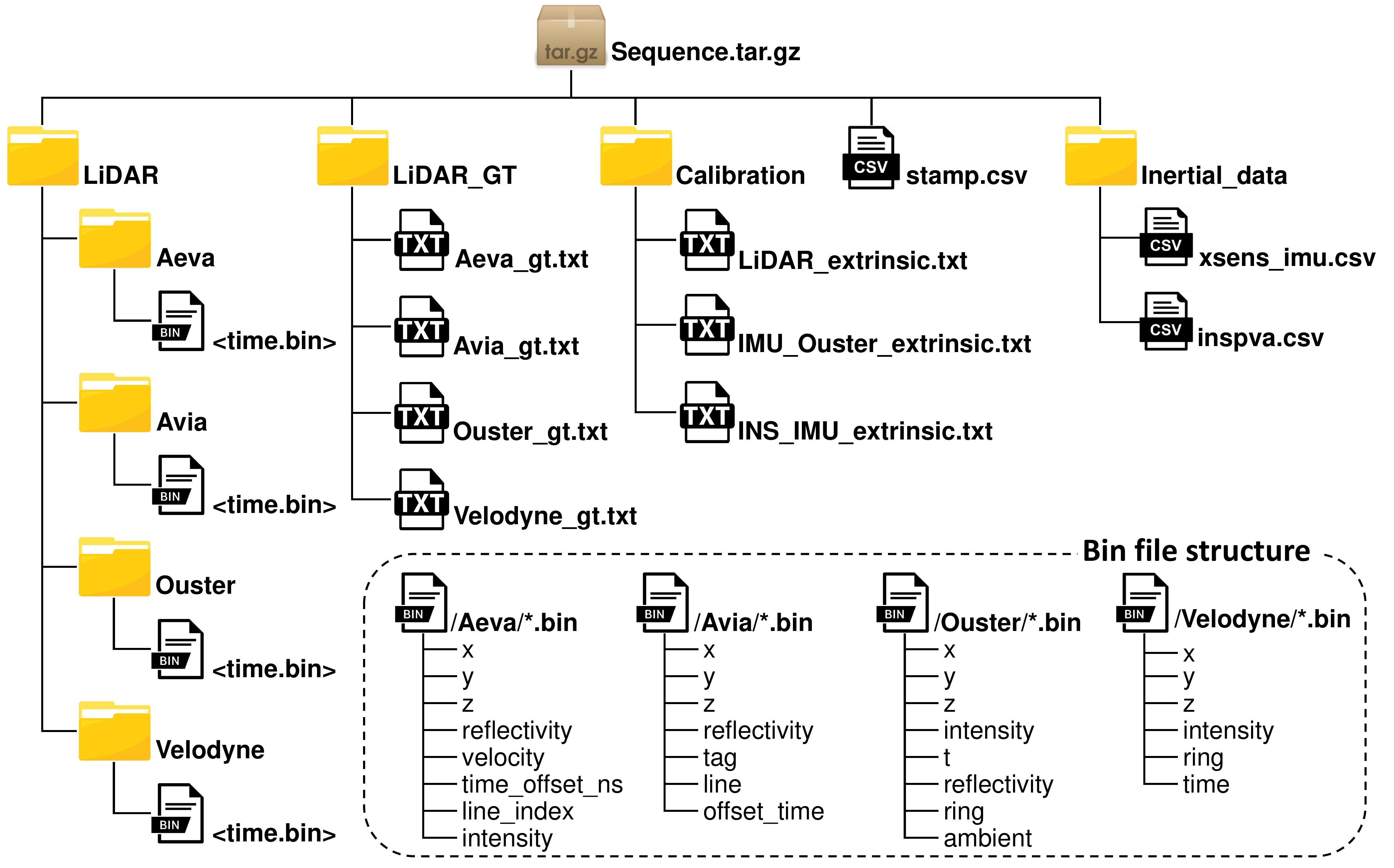}
    \caption{File structure of the HeLiPR dataset, illustrating the organization of LiDAR scans, ground truths, calibration, and inertial sensor measurements for each sequence}
    \label{fig:FileSystem}
    \vspace{-5mm}
\end{figure}

\section{Description of HeLiPR Dataset}
\subsection{Data Format}
We offer sensor-specific data as individual files in diverse formats to optimize dataset management and facilitate access to each file and frame. Furthermore, we supply a file player based on \ac{ROS}, tasked with reading and publishing these files into ROS topics, ensuring seamless accessibility for pre-existing place recognition and tasks like \ac{SLAM}. The file structure of the HeLiPR dataset is delineated in \figref{fig:FileSystem}. The acquisition time of all measurements is stored in \texttt{stamp.csv}, and detailed descriptions of the data are presented subsequently.
\subsubsection{Multiple LiDARs Data}
Individual LiDAR scans are stored as binary files in the \texttt{LiDAR/Sensor\_name}. These files, identified as \texttt{<time.bin>}, encompass common channels such as $(x, y, z)$, time offset, and ring (or line) index. We illustrate the array of unique channels for each LiDAR and the order of their storage in \figref{fig:FileSystem}.
\subsubsection{INS Data}
All INS data are stored in the \texttt{inspva.csv}. This file includes time, latitude, longitude, height, north\_velocity, east\_velocity, up\_velocity, roll, pitch, azimuth, and data\_status, organized in this order. Each value adheres to the \ac{ENU} coordinate system, with the azimuth being determined by a left-handed rotation around the z-axis, in degrees, and clockwise from north.
\subsubsection{IMU Data}
The complete set of IMU data is contained within the \texttt{xsens\_imu.csv} file. Sequentially, this file encompasses time, quaternion $(x, y, z, w)$, Euler angles $(x, y, z)$, gyroscope $(x, y, z)$, acceleration $(x, y, z)$, and magnetic field $(x, y, z)$.
\subsubsection{Calibration and Ground truth Data}
The results of extrinsic calibration are saved in the \texttt{Calibration}. Additionally, we derive the individual LiDAR ground truth based on INS, LiDAR acquisition time and calibration parameters. Within the \texttt{LiDAR\_GT}, the ground truth for each LiDAR is recorded, incorporating scan time, position $(x, y, z)$, and quaternion $(x, y, z, w)$. The procedure for generating this file is discussed in \secref{sec:groundtruth}.

\begin{figure}[!t]
    \centering
    \includegraphics[width=\columnwidth]{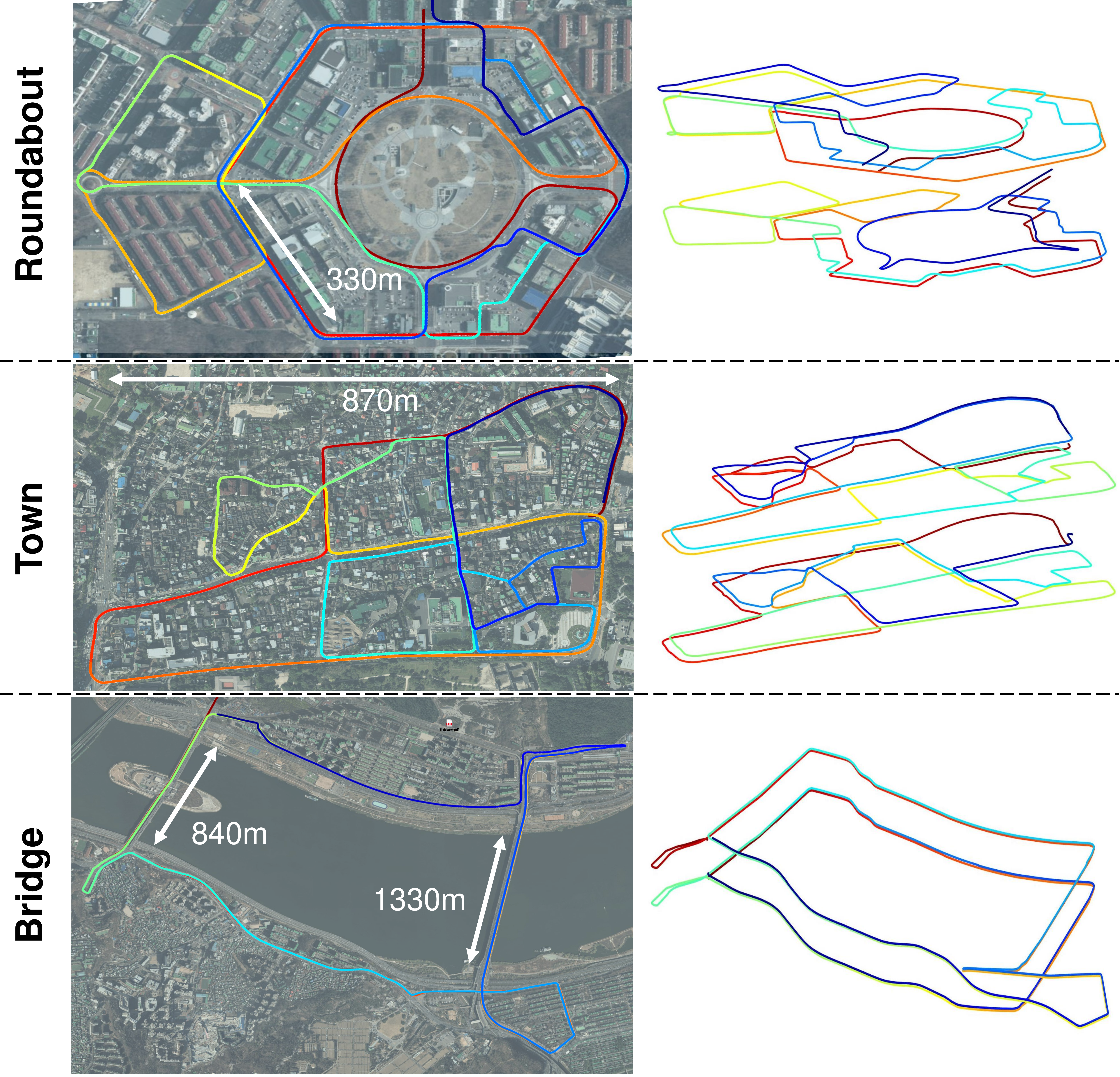}
    \caption{INS-based trajectories for sequences \texttt{01}, \texttt{02}, and \texttt{03}. The left shows trajectories on aerial images for \texttt{01}, while the right visualizes \texttt{02} (bottom) and \texttt{03} (top) with a color gradient. Notably, red indicates the start point, while blue designates the endpoint.}
    \label{fig:sequenceTraj}
    \vspace{-5mm}
\end{figure}

\begin{figure*}[!t]  
    \centering
    \centering
    \subfigure[\texttt{Roundabout}]
    {   \includegraphics[height=0.24\textwidth]{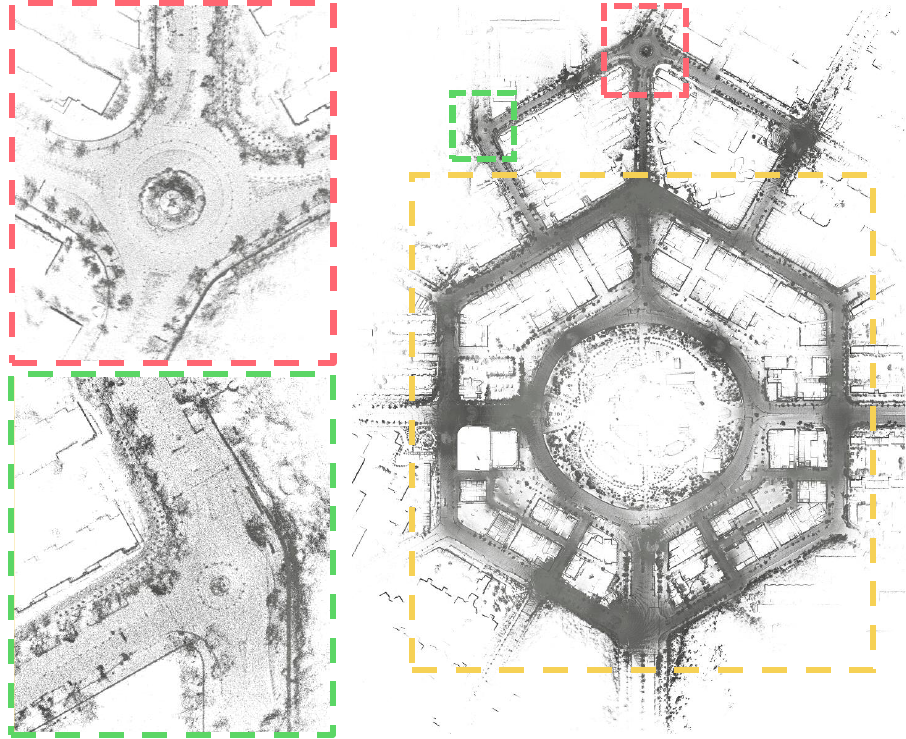}
    \label{fig:Dataset_a}
    }
    \subfigure[\texttt{Town}]
    {
    \includegraphics[height=0.24\textwidth]{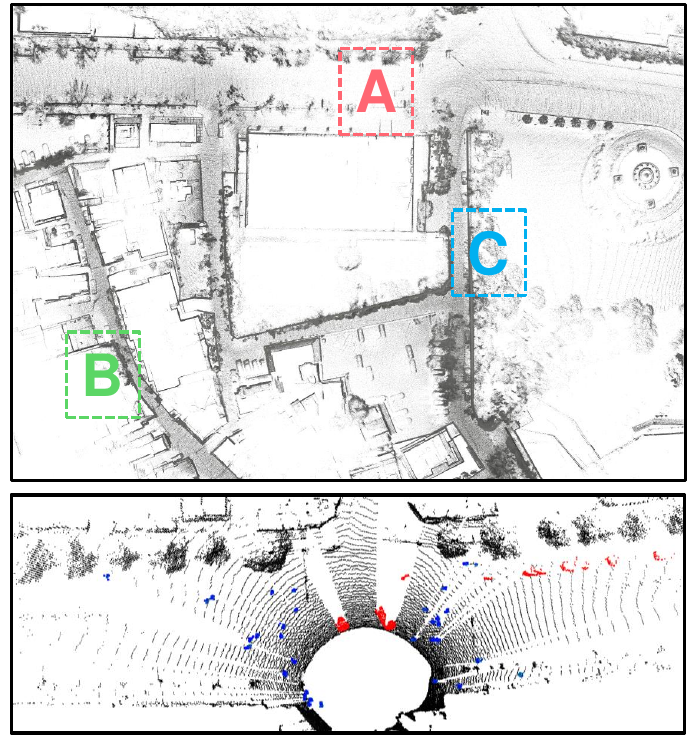}
    \label{fig:Dataset_b}
    }
    \subfigure[\texttt{Bridge}]
    {
    \includegraphics[height=0.24\textwidth]{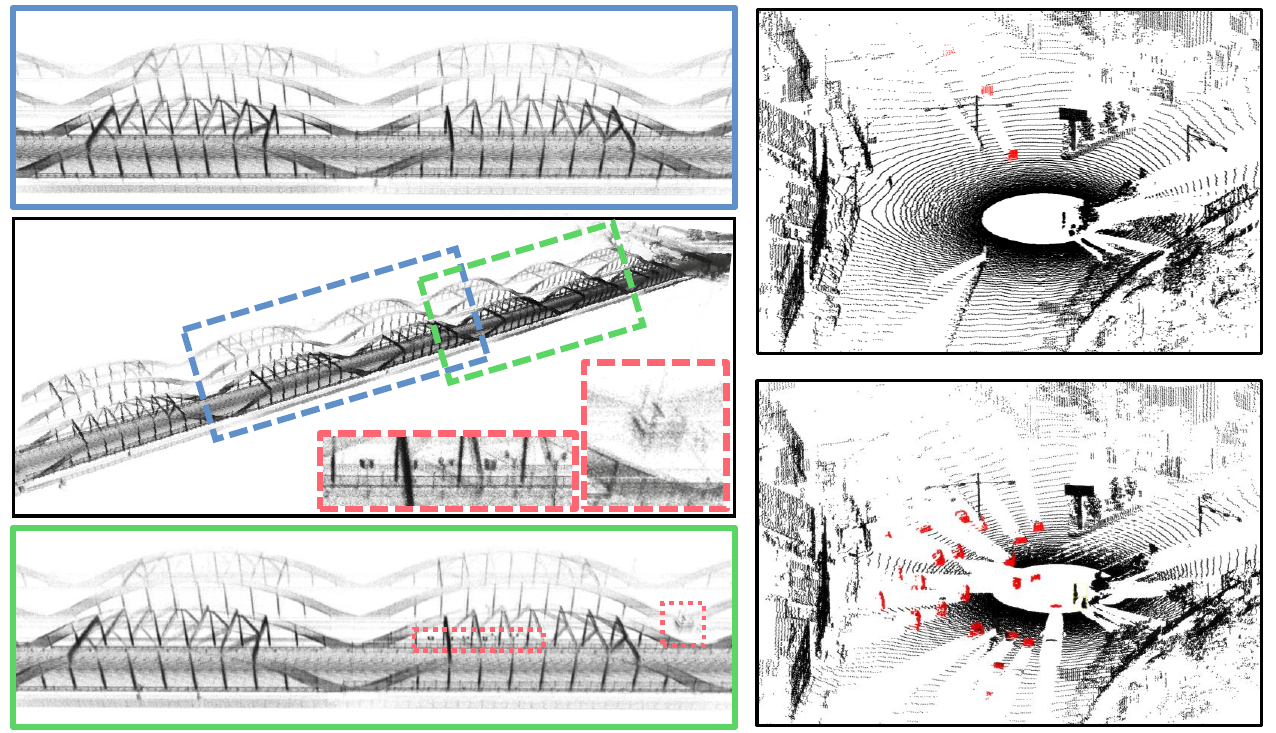}
    \label{fig:Dataset_c}
    }    
    \vspace{-4mm}
    \caption{Characteristics of sequences. \subref{fig:Dataset_a} \texttt{Roundabout} with three distinct roundabouts, each highlighted by a colored box. \subref{fig:Dataset_b} \texttt{Town} showcasing a mix of narrow alleys and wider boulevards with width indications for areas A, B, and C, which are measured approximately 15 m, 3 m and 5 m in width. Dynamic entities, such as pedestrians (blue) and vehicles (Red), are marked. \subref{fig:Dataset_c} \texttt{Bridge} emphasizing the challenge of scene similarities within the sequence. The portions of the bridge highlighted in blue and green boxes appear remarkably similar but display subtle differences, as indicated in the red box. Additionally, the presence of dynamic objects (red) and variations in numbers between \texttt{Bridge01} (upper) and \texttt{Bridge02} (lower) pose quite challenges for place recognition.}
    \label{fig:Dataset}
    \vspace{-5mm}
\end{figure*}

\subsection{Sequence Explanation}
In the HeLiPR dataset, we present three distinct places, namely \texttt{Roundabout}, \texttt{Town}, and \texttt{Bridge}. These places are meticulously acquired through three repetitions denoted as \texttt{01-03}, with a two-week interval between each acquisition. The deliberate interval introduces temporal changes to enable inter-session place recognition. Furthermore, this temporal variation encompasses both night and day environments, leading to notable variations in the presence of dynamic objects throughout the sequences. It also allows for spatial variations, such as lane changes or reversing directions, when capturing data at the same location but on different paths. Detailed information, including acquisition time, duration, and distance, can be found in the \tabref{tab:data_description}. Each sequence showcases unique environmental characteristics and introduces novel challenges in inter-LiDAR place recognition. We focus on enhancing both intra-session and inter-session loop closure candidates, with the primary objective of generating an abundant set of queries for place recognition. All sequences' trajectory and characteristics are represented in \figref{fig:sequenceTraj} and \figref{fig:Dataset}.

\textbf{(\textit{i}) \texttt{Roundabout01-03}: } 
\texttt{Roundabout} stands out as the most formal environment for place recognition among all the sequences. Tall buildings and wide roads enrich the dataset with abundant features that aid in place recognition. As its name suggests, it consists of three roundabouts: one large and two of a comparatively smaller size, as shown in \figref{fig:Dataset_a}. The presence of a large roundabout and an outer hexagon design ensures easy revisiting of previously encountered locations. Moreover, the interconnected layout of roads and alleys within both the roundabout and hexagon facilitate seamless movement, enabling access to various exits from any entrance. These spatial features provide diverse candidates for place recognition and rotational variations not typically encountered in regular road scenarios.

\textbf{(\textit{ii}) \texttt{Town01-03}: } 
\texttt{Town} presents a wide road environment in the center of the route, facilitating efficient scanning of buildings and structures. This characteristic shares similarities with the \texttt{Roundabout} sequence. However, in \texttt{Town}, the buildings are relatively short compared to those in \texttt{Roundabout}, and narrow alleys are more frequent. These alleys pose challenges in utilizing the wide sensing capabilities of LiDAR, creating a situation akin to indoor place recognition. These challenges can be found in \figref{fig:Dataset_b}. Furthermore, the presence of diverse dynamic objects in narrow alleys contributes to a lower proportion of static objects in the scene. Spinning LiDAR systems effectively address these limitations with their expansive FOV. In contrast, solid-state LiDAR systems inherently have a narrower FOV, potentially leading to a significantly reduced detection of static objects. These environment-related disparities add another layer of complexity to place recognition, demanding sophisticated approaches to handle such spatial variations effectively.

\textbf{(\textit{iii}) \texttt{Bridge01-04}: }
\texttt{Bridge} consists of a total of 2 laps, covering two bridges with lengths of 1.3 \unit{km} and 0.8 \unit{km}, respectively. It should be noted that \texttt{Bridge01} and \texttt{Bridge04} differ from \texttt{Bridge02} and \texttt{Bridge03}, with the former being driven using a reverse trajectory. This place introduces a significant challenge in place recognition due to the consecutive appearance of similar scenes in most areas. The appearances with small differences represented in \figref{fig:Dataset_c} can lead to result in many false positives. Moreover, numerous dynamic objects add complexity to place recognition, particularly depending on the sequence index. Notably, \texttt{Bridge02} and \texttt{Bridge04} display a relatively slow speed distribution attributed to the high density of dynamic objects. The Aeries II, which is a \ac{FMCW} LiDAR, allows for measuring the radial velocity of a point, enabling the detection of certain dynamic objects. This capability opens up the potential for developing a novel place recognition based on this specific sequence and leveraging the unique features of the Aeries II.

\begin{figure}[!t]
    \centering
    \includegraphics[width=\columnwidth]{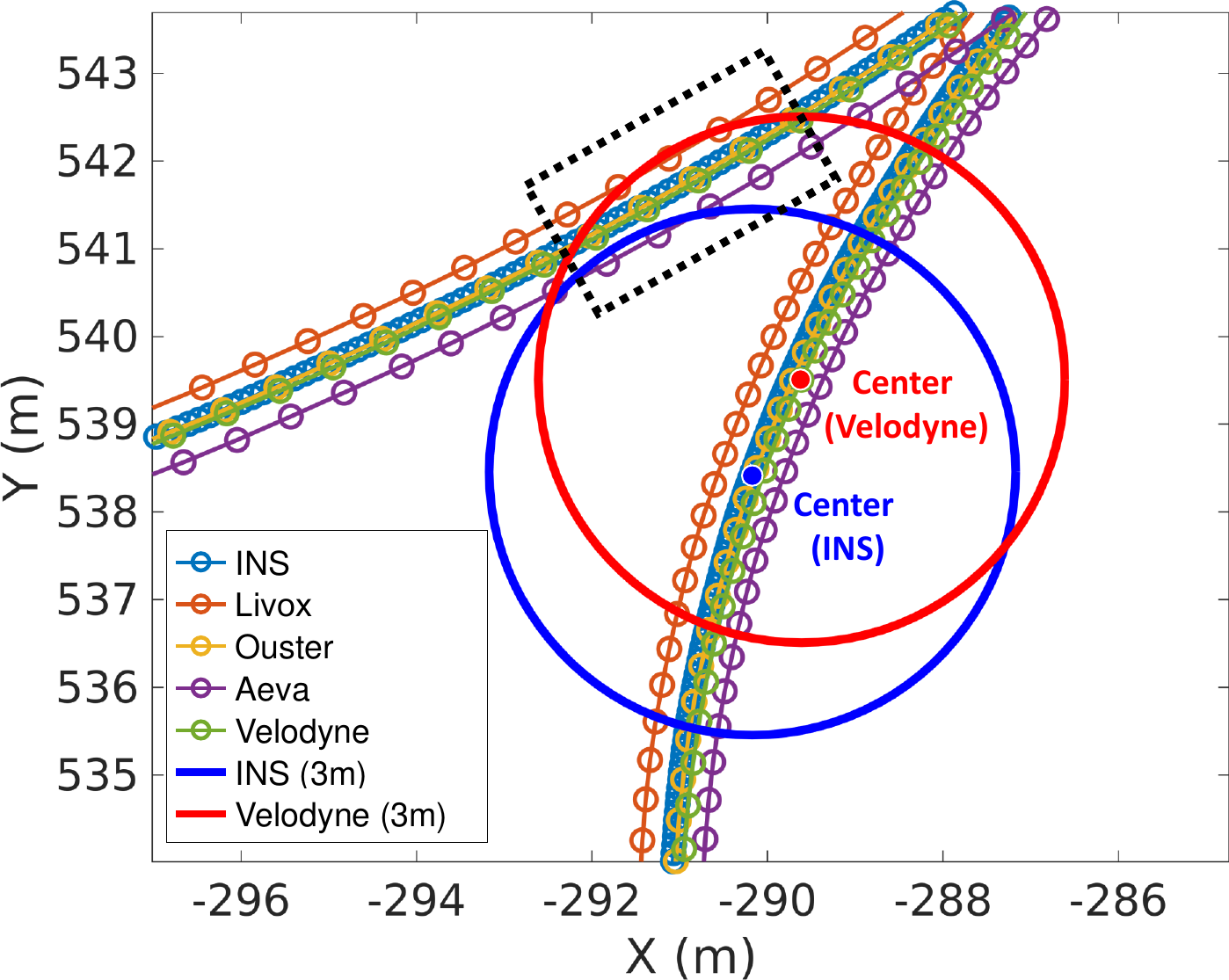}
    \vspace{-5mm}
    \caption{Partial individual ground truth for \texttt{Roundabout01}. Using individual ground truth with Velodyne as a reference, true positive place recognition candidates are observed for all LiDARs within a 3 m radius of other paths (black box). However, exact matches with the INS location (within a 5ms discrepancy) are only found in Aeva.}
    \label{fig:Groundtruth}
    \vspace{-5mm}
\end{figure}

\subsection{Estimating Individual LiDAR Ground Truth with B-spline Interpolation}
\label{sec:groundtruth}
In place recognition, the positions of the query and the candidate must be determined, necessitating the precise positions of the LiDAR at each timestamp. Solutions such as GPS, INS, or SLAM can be employed to determine the trajectory of the system, with the resulting trajectory treated as the ground truth. However, considering that this ground truth represents the location of a reference point at specific timestamps, it is essential to account for spatiotemporal variances when expressing precise positions for all LiDARs. Extrinsic calibration, defining the spatial relationship, introduces differences between the LiDAR and the ground truth. Although all LiDARs capture scans under \ac{UTC} via \ac{PTP}, their individual acquisition times vary, leading to temporal discrepancies. As a result, timestamps from each LiDAR do not align with the ground truth time. The ground truths for each LiDAR and the single source are depicted in Figure \ref{fig:Groundtruth}, emphasizing the necessity of obtaining ground truths for all LiDARs by considering spatiotemporal variances rather than relying on a single ground truth. The figure also demonstrates that assuming Velodyne is at the same position as the ground truth from a single source results in only the Aeva set being identified as true positive. However, accounting for discrepancies allows different LiDAR sets to be recognized as true positives. This underscores the importance of considering these variances for more comprehensive and accurate place recognition. Therefore, incorporating such discrepancies is beneficial when leveraging multiple LiDARs, providing a more practical approach than relying on a single ground truth.

For the HeLiPR dataset, INS serves as ground truth. While \ac{INS} operates at a frequency of 50 Hz, which is five times faster than the LiDAR, timestamps of \ac{INS} are not synchronous with timestamp of LiDARs. To mitigate this issue, we calculate the location of LiDARs at specific timestamp using B-spline interpolation \citep{mueggler2018continuous} based on \ac{INS} position, as relying solely on linear approximation may lead to imprecision. We choose B-spline interpolation for its compatibility with the inherently smooth characteristics of real vehicle trajectories. Given that the $k^{th}$ scan of a specific LiDAR, denoted as $L_k$, is acquired at time $t_k$, this position, $\textbf{T}^N_W = (\textbf{R}^N_W, \textbf{t}^N_W)$ can be determined by leveraging four nearby \ac{INS} measurements as control points. Nevertheless, our primary interest lies in determining the location of $\textbf{T}^L_W = (\textbf{R}^L_W, \textbf{t}^L_W)$ in the world coordinates of $L$. It can be calculated as the multiplication with $\textbf{T}^N_W$, IMU-LiDAR $(\textbf{T}^L_I)$ and INS-IMU $(\textbf{T}^I_N)$ extrinsic calibration. 
To achieve user convenience, we standardize \( \textbf{T}^L_W \) using the \ac{UTM} coordinate system, opting not to employ latitude or longitude. This decision simplifies plotting processes, and these coordinates streamline the direct comparison of trajectories with each other and the MulRan dataset.

In summary, our approach goes beyond simply utilizing multiple LiDARs for place recognition. By embracing the individual trajectories and specific ground truths of each LiDAR, we enable a more comprehensive evaluation of spatiotemporal variations. Our diverse ground truths significantly improve the accuracy and reliability of place recognition outcomes over relying solely on a single ground truth.

\subsection{HeLiPR Dataset: Long-Term Place Recognition in Tandem with MulRan}

\begin{table*}[!t]
\caption{The Description for Additional Sequences.}
\label{tab:MulRan}
\resizebox{\textwidth}{!}{%
\begin{tabular}{c|cccccccccccc} \toprule
\multirow{3}{*}{Sequence Name}  & \multicolumn{12}{c}{Sequence Index}                                                                          \\
                               &                     \multicolumn{4}{c}{04}                   & \multicolumn{4}{c}{05}          & \multicolumn{4}{c}{06}  
                               \\
                               &  Date & Time &                Duration & Distance                &  Date & Time & Duration & Distance  & Date & Time& Duration & Distance             \\ \midrule
\texttt{KAIST}    &   2023-08-31     & Midnight   & 1261 \unit{s}      & \multicolumn{1}{c|}{6348 \unit{m}}& 2023-08-31 & Daytime  &1248 \unit{s}          & \multicolumn{1}{c|}{6878 \unit{m}}& 2024-01-16 & Night &1215 \unit{s}      & \multicolumn{1}{c}{6661 \unit{m}}   \\
\texttt{DCC}  &  2023-08-31  & Midnight &786 \unit{s}        & \multicolumn{1}{c|}{5506 \unit{m}} & 2023-08-31 & Daytime &1081 \unit{s}        & \multicolumn{1}{c|}{5309 \unit{m}}& 2024-01-16 &  Night &1074 \unit{s}      & \multicolumn{1}{c}{4648 \unit{m}}     \\
\texttt{Riverside} &  2023-08-31 &  Midnight&612 \unit{s}        & \multicolumn{1}{c|}{6523 \unit{m}} & 2023-08-31 &  Daytime &855 \unit{s}         & \multicolumn{1}{c|}{6394 \unit{m}} & 2024-01-16 & Night & 1195 \unit{s}      & \multicolumn{1}{c}{7219 \unit{m}}     \\ \bottomrule
\end{tabular}%
}
\vspace{-3mm}
\end{table*}

\begin{figure}[!t]
    \centering
    \includegraphics[width=\columnwidth]{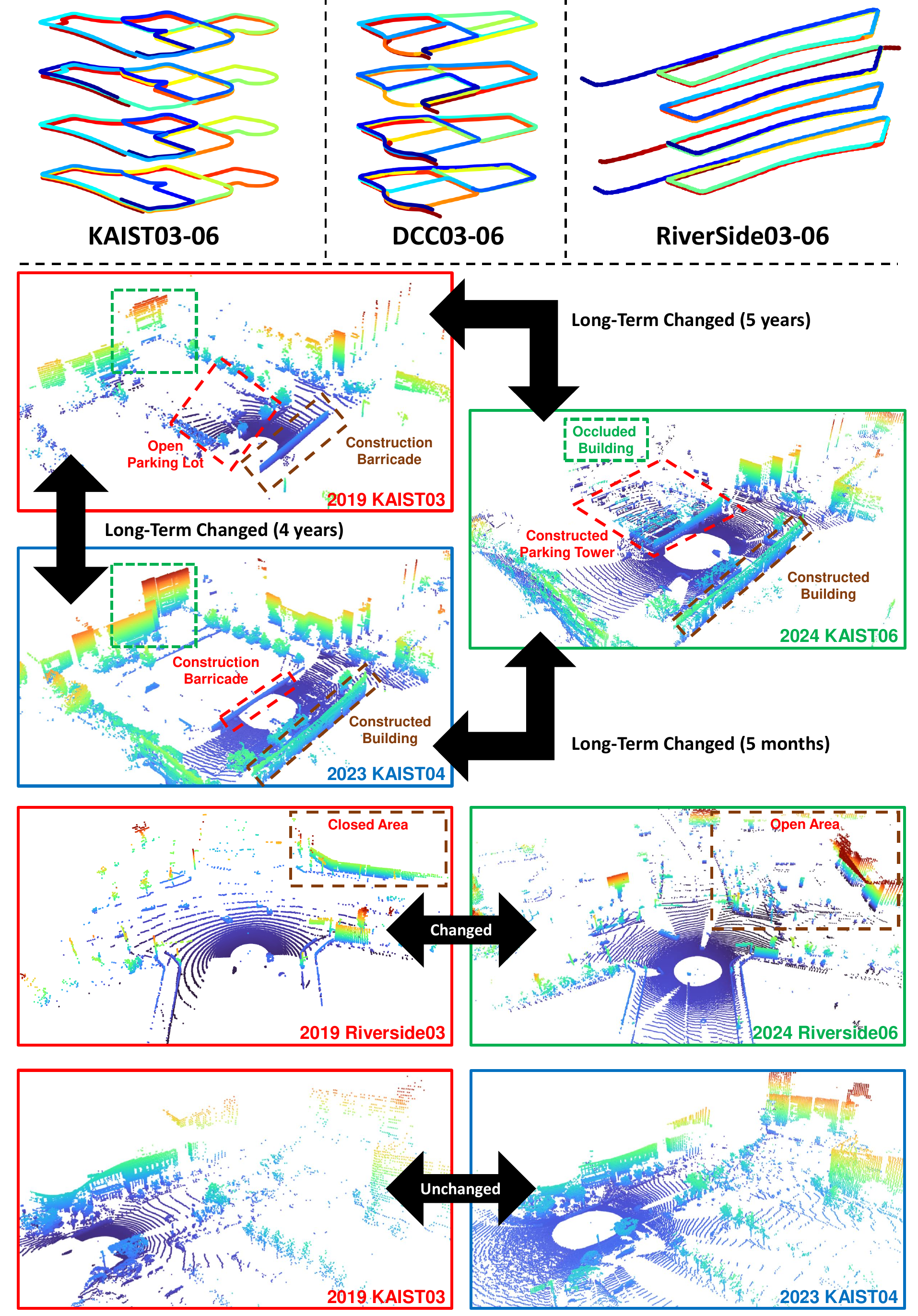}
    \caption{(Top row) Trajectories from GPS and INS data with three paths for each location, as \figref{fig:sequenceTraj}. (Bottom row) shows environmental changes in the order \texttt{03}, \texttt{04}, and \texttt{06}. The \texttt{KAIST} depicts gradual scene changes at a similar location. \texttt{Riverside} highlights the distinct scan differences caused by different LiDARs and shows drastic changes between open and closed areas. The last row shows similar scans even after a long period.}
    \label{fig:Mulran}
    \vspace{-7mm}
\end{figure}

The HeLiPR dataset not only encompasses the sequences of \texttt{Roundabout}, \texttt{Town}, and \texttt{Bridge}, but also integrates sequences from the MulRan dataset, including \texttt{KAIST}, \texttt{DCC}, and \texttt{Rivierside}, catering to long-term place recognition. Contrary to the MulRan dataset captured with OS1-64, the HeLiPR dataset employs the LiDARs as mentioned earlier. This introduces the potential for heterogeneous LiDAR place recognition. Furthermore, a temporal variance spanning approximately four years offers a novel challenge: long-term place recognition.

The Long-term place recognition challenge within the HeLiPR dataset includes sequences \texttt{04-06}. The \texttt{04} sequence, captured at midnight, exhibits an almost complete absence of dynamic objects. IMU measurements are not included in this sequence but are not crucial for place recognition. In contrast, the \texttt{05} sequence, captured during the daytime, features diverse dynamic objects resembling the existing MulRan sequences. Unlike \texttt{04} and \texttt{05}, \texttt{06} was captured four months later. This period, shorter than the four-year gap with MulRan, reduces its complexity as a long-term place recognition challenge. Nonetheless, the temporal differences between \texttt{04}, \texttt{05}, and \texttt{06} offer a distinct perspective, capturing scene changes over a more moderate term. A comprehensive overview of each sequence can be found in \tabref{tab:MulRan} and \figref{fig:Mulran}. Thanks to GPS and INS, the trajectories from MulRan and HeLiPR align seamlessly despite being captured at distinct times, facilitating place recognition. As observed in \figref{fig:Mulran}, certain areas show partial modifications. From \texttt{KAIST03} to \texttt{KAIST06}, for instance, the construction of parking lots and buildings leads to significant scene changes. These new structures alter the landscape and occlude the LiDAR, causing scenes to appear different even in the same location. Such changes are crucial for both place recognition and change detection, as well as for map maintenance. Furthermore, in \texttt{Riverside03}, the scanning coverage is more limited compared to the current platform, with obstructions like a construction barricade blocking the upper right view from the vehicle. This limitation emphasizes the need to carefully compare similar areas and reduce reliance on regions with substantial scene changes for long-term place recognition in \texttt{Riverside06}. The challenge lies in accurately identifying revisits to specific locations, particularly in scenarios where parts of the environment have changed. In such cases, it becomes crucial not only to recognize these changes but also to possess the capability to compare current scenes with previous ones by focusing on elements that remain unchanged. This situation also prompts further exploration into the extent of scene changes that can be accommodated in place recognition, questioning whether significantly altered locations can still be recognized as the same for revisiting purposes.

\section{Benchmark Results with HeLiPR Dataset}

This section presents an exhaustive analysis of state-of-the-art place recognition using the HeLiPR dataset. The comprehensive evaluation aims to identify the capabilities of state-of-the-art place recognition and emphasize the inherent need and importance of having datasets like HeLiPR to advance the field further.
To assess the performance of the methodologies, which include Scan Context (SC) \citep{kim2021scan}, RING++ \citep{9981308}, BTC \citep{10388464} and LoGG3D-Net \citep{vid2022logg3d}, we employ three evaluation metrics: the Precision-Recall curve (PR-curve), the \ac{AUC} score and R@1\%.

\textbf{Precision-Recall Curve (PR-Curve)}: This curve is an illustrative representation of a precision versus its recall. It provides a comprehensive visualization of the performance across different threshold levels. Mathematically, it can be expressed by the following equations:
\begin{equation}
\text{Precision} = \frac{\text{TP}}{\text{TP} + \text{FP}}, \;\;
\text{Recall} = \frac{\text{TP}}{\text{TP} + \text{FN}}
\end{equation}
where TP denotes true positives, FP represents false positives, and FN signifies false negatives.

\textbf{Area Under the Curve (AUC)}: This metric evaluates the overall performance of a place recognition. The AUC score provides a singular scalar value summarizing the entire PR curve. A perfect recognition method would achieve an AUC of 1.0, indicating flawless recognition, while a score closer to 0.5 might indicate a method performing no better than random guessing.

\textbf{R@N(\%)}: R@N(\%) evaluates the recall within the top $N$ or $N$ percentage of results, which is calculated by determining if the true positive pairs are within the $N$ or $N$ percentage of the closest matches in the database. This metric provides insight into how effectively the algorithm identifies the most accurate matches from a larger pool of candidates, offering a more detailed perspective on its precision in high-relevance scenarios. For the benchmark results, we select $N=1$ and represent both R@N and R@N\%.

\begin{table*}[!t]
\centering
\caption{AUC score and R@N for inter-LiDAR and inter-session place recognition}
\label{tab:AUC}
\resizebox{\textwidth}{!}{%
\begin{tabular}{c|c|cccccc|ccccccccc} \toprule
\multirow{3}{*}{Sequences} & \multirow{3}{*}{Method} & \multicolumn{6}{c|}{Identical LiDARs} & \multicolumn{9}{c}{Heterogeneous LiDARs} \\
                                 &        & \multicolumn{3}{c}{O-O}                  & \multicolumn{3}{c|}{A-A}                   & \multicolumn{3}{c}{O-V}                  & \multicolumn{3}{c}{O-A}                  & \multicolumn{3}{c}{O-L}                  \\ 
                                 &  &  AUC & R@1 & R@1\% &  AUC & R@1 & R@1\% &  AUC & R@1 & R@1\% &  AUC & R@1 & R@1\% &  AUC & R@1 & R@1\%  \\
                                 \midrule[1pt]
\multirow{4}{*}{\texttt{Roundabout01-03}} 
                                 & SC     & 0.942       &0.880&0.947& 0.786       &0.648&0.681& 0.119       &0.077&0.130& 0.007       &0.008&0.018& 0.017       &0.018&0.030\\
                                 & BTC    & 0.669       &0.604&0.946& 0.702       &0.575&0.672& 0.302       &0.323&0.789& 0.113       &0.114&0.359& 0.088       &0.100&0.413  \\
                                 & RING++ & 0.950       &0.926&0.993& 0.809       &0.640&0.704& 0.067       &0.069&0.251& 0.003       &0.004&0.052& 0.003       &0.004&0.047\\ 
                                 & LoGG3D & 0.766 & 0.552 & 0.639 & 0.746 & 0.538 & 0.629 & 0.531 & 0.442 & 0.612 & 0.641 & 0.474 & 0.605 & 0.392 & 0.333 & 0.522 \\ 
                                 \midrule
\multirow{4}{*}{\texttt{Town01-03}}       
                                 & SC     & 0.957       &0.826&0.918& 0.811       &0.601&0.719& 0.312       &0.174&0.219& 0.024       &0.020&0.057& 0.095       &0.068&0.133\\
                                 & BTC    & 0.625       &0.463&0.650& 0.685       &0.482&0.569& 0.343       &0.291&0.467& 0.149       &0.136&0.377& 0.153       &0.149&0.403 \\
                                 & RING++ & 0.965       &0.935&0.991& 0.919       &0.718&0.778& 0.098       &    0.083    &0.252& 0.062       &0.009&0.071& 0.004       &0.005&0.054\\
                                 & LoGG3D & 0.829 & 0.607 & 0.705 & 0.779 & 0.553 & 0.687 & 0.448 & 0.354 & 0.575 & 0.611 & 0.452 & 0.634 & 0.267 & 0.258 & 0.424 \\ \midrule
\multirow{4}{*}{\texttt{Bridge02-03}}     
                                 & SC     & 0.713       &0.786&0.948& 0.666       &0.712&0.925& 0.108       &0.091&0.183& 0.019       &0.018&0.035& 0.012       &0.013&0.041\\
                                 & BTC    & 0.447       &0.422&0.676& 0.396       &0.423&0.707& 0.187       &0.226&0.463& 0.060       &0.079&0.305& 0.071       &0.092&0.370  \\
                                 & RING++ & 0.868       &0.855&0.995& 0.802       &0.800&0.993& 0.037       &0.049&0.238& 0.005       &0.018&0.035& 0.005       &0.013&0.041\\
                                 & LoGG3D & 0.692 & 0.670  & 0.903 & 0.612  & 0.599 & 0.890  & 0.347 & 0.389 & 0.778 & 0.486 & 0.518 & 0.851 & 0.263 & 0.303 & 0.597 \\
                                  \bottomrule
\end{tabular}%
}
\tiny
\; \\
\footnotesize 
\raggedright
- Symbol denotes LiDARs. (O: Ouster, A: Aeva, L: Livox, V: Velodyne)
\vspace{-5mm}
\end{table*}

\begin{figure}[!t]
    \centering
    \subfigure[PR-Curve for Identical LiDAR (\texttt{Bridge})]
    {%
    \includegraphics[width=\columnwidth]{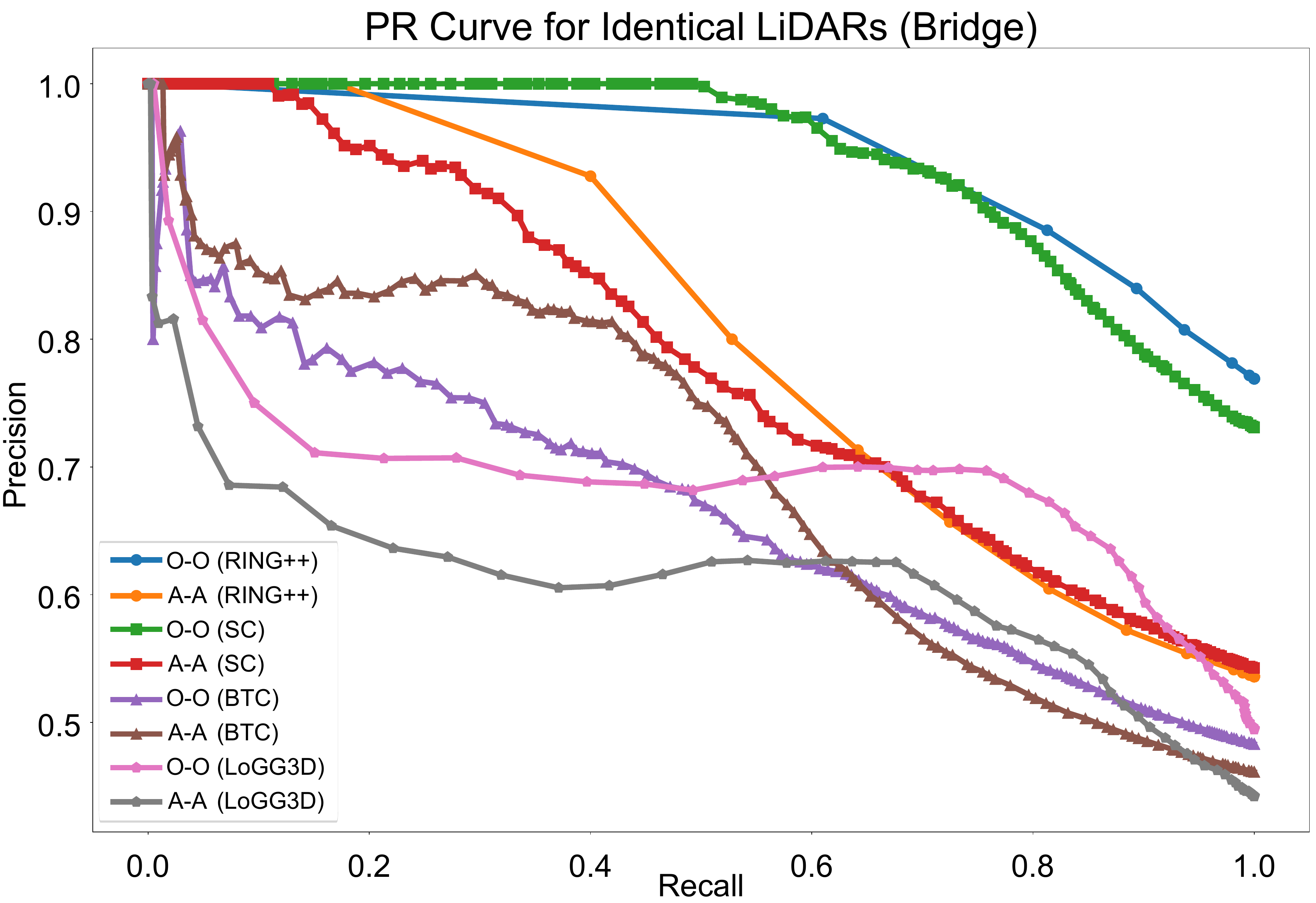}
    \label{fig:pr_curve_a}
    } 
    \subfigure[PR-Curve for Heterogeneous LiDAR (\texttt{Bridge})]
    {
    \includegraphics[width=\columnwidth]{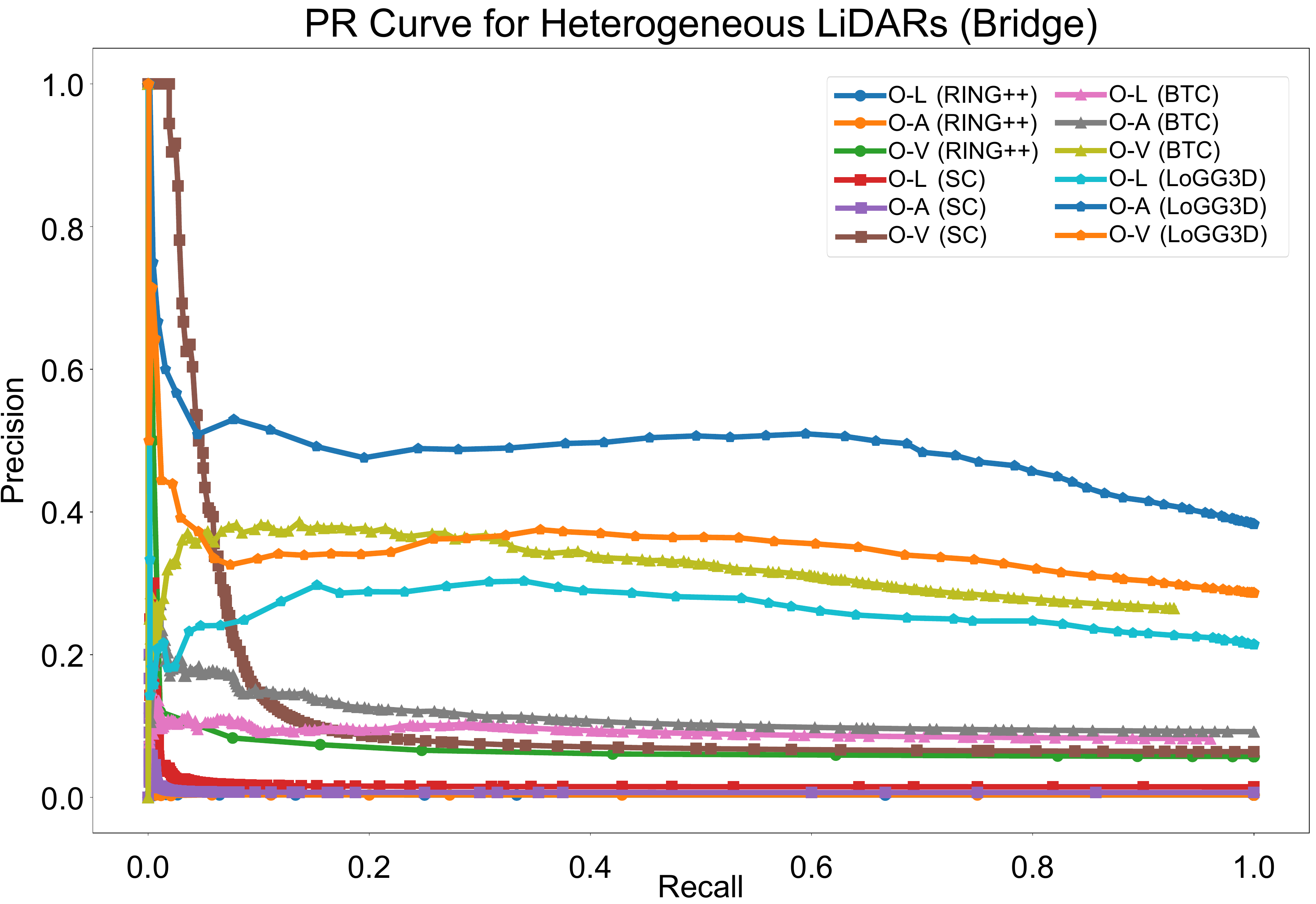}
    \label{fig:pr_curve_b}
    }
    \vspace{-2mm}
    \caption{PR-Curve of inter-session place recognition between \texttt{Bridge02} and \texttt{Bridge03}. \subref{fig:pr_curve_a} depicts the results when identical LiDAR systems are compared, illustrating the performance when different types of LiDARs each match with an identical LiDAR. \subref{fig:pr_curve_b} showcases the outcomes when heterogeneous LiDARs are used.}
    \label{fig:pr_curve}
    \vspace{-7mm}
\end{figure}

For the evaluation, the methodology entails sampling query scans at 10 m intervals and target scans at 5 m intervals. Successful place recognition is defined by identifying a candidate within a 7.5 m, termed a true positive. All scans have been undistorted and are configured with a maximum range of 100 m for descriptor extraction. For methods other than BTC, only the scans from Livox Avia are grouped into sets of 20 due to their sparse point distribution. However, when evaluating BTC, similar to the original research, we accumulate 20 scans for every type of LiDAR. This approach differentiates it from other methods that typically use a single scan.
LoGG3D-Net is selected for the deep learning-based place recognition method since LoGG3D-Net provides benchmark results of a mean maximum F1 score on the KITTI and MulRan datasets. The parameters and strategies for training can be found in our project page.

We evaluate each method using three inter-session pairs: \texttt{Roundabout01-03}, \texttt{Town01-03}, and \texttt{Bridge02-03}. Using Ouster as the reference database, we employ various LiDARs as queries to identify the corresponding Ouster candidates. To assess the influence of FOV, we also perform evaluations using the same LiDAR type. As detailed in \tabref{tab:AUC}, we evaluate the four methods across three environments using five LiDAR pairings. This experimental configuration underscores the expansive scenarios of place recognition that the HeLiPR dataset can facilitate.

From \figref{fig:pr_curve} and \tabref{tab:AUC}, we can discern several insights about the performance of place recognition. In terms of the dataset perspective, most methods tend to exhibit superior performance in \texttt{Roundabout} and \texttt{Town} compared to \texttt{Bridge}, likely due to the distribution of structures and inherent challenges highlighted in \figref{fig:Dataset}. Furthermore, for the model-based method, spinning LiDAR generally achieves better results with a higher AUC score and R@N than solid-state LiDAR. This is anticipated, given that spinning LiDARs have a large FOV. This expansive FOV equips them to adeptly handle place recognition from varied directions, including in scenarios like reverse visiting or navigating intersections.

In inter-LiDAR place recognition, we observed specific challenges associated with different LiDAR pairings: Ouster and Velodyne show resolution differences, Ouster and Aeva differ in horizontal FOV, and Ouster and Livox vary in both horizontal FOV and scanning patterns. These differences significantly affect the performance of model-based methods like RING++ and Scan Context, which are sensitive to LiDAR FOV variations as these descriptors are constructed with different FOVs in inter-LiDAR place recognition, a notable performance dip is observed in inter-LiDAR place recognition.

Contrary to other methods, BTC stands out in inter-LiDAR place recognition, particularly excelling in R@1\% performance compared to other model-based approaches; however, its AUC and R@1 scores fall short of expectations. This discrepancy arises from the nearest target not having the largest similarity, meaning that the similarities between the query and target are not always aligned with the distance. Furthermore, there are instances where the overlap is significant, but the distance exceeds 7.5m, resulting in a false negative. Consequently, while the AUC and R@1 might seem underwhelming since they only utilize the best candidates, the R@1\% shows a improvement due to the enhanced possibility of identifying compatible matches among the total candidates. Unlike traditional place recognition for spinning LiDAR, place recognition of heterogeneous LiDAR may have a smaller overlap depending on the FOV or maximum range, even if the distance between query and target is close. Therefore, it is also proposed to distinguish based on overlap rather than distance as a criterion for determining true positives.

For the learning-based method, LoGG3D-Net, a slight underperformance is noted in identical LiDAR place recognition compared to model-based methods. However, it performs better inter-LiDAR place recognition, benefitting from training with super sequences that enhance its ability to distinguish between heterogeneous LiDARs. This is evident in its AUC score and R@N, with robust results in Ouster-Aeva comparisons. The similarity in the number of vertical channels between Ouster and Aeva likely contributes to better local feature aggregation. In contrast, Ouster and Livox exhibit lower scores, primarily because of the significant differences between their LiDAR characteristics, especially in comparison to Ouster and Velodyne. This emphasizes the sensitivity of LiDAR performance to resolution and FOV, with dual degradation occurring when FOV and scanning patterns differ. While the learning-based method shows promise for inter-LiDAR place recognition, effectively handling heterogeneous LiDARs remains challenging. 

In summary, model-based methods excel when using identical LiDARs but fall short in inter-LiDAR scenarios. On the other hand, learning-based methods maintain consistent performance across various LiDAR combinations but only achieve partially satisfactory results. Recent learning-based methods try to perform place recognition with FOV variations \citep{kong2020semantic, vidanapathirana2021locus}, it do not present reliable performances since FOV variances are relatively smaller than the difference between solid state and spinning LiDAR. All approaches currently need to be revised to achieve the desired performance levels. This analysis highlights the need for focused research in heterogeneous inter-LiDAR place recognition, with the HeLiPR dataset serving as a valuable resource for such investigations.
\section{Conclusion}
The HeLiPR dataset stands as a comprehensive resource that has been meticulously curated to showcase the remaining challenges of place recognition. It encompasses a broad spectrum of data from varied environments, including \texttt{Roundabout}, \texttt{Town}, and \texttt{Bridge}. One of the unique attributes of this dataset is its collection method; by introducing intentional time intervals and capturing data along diverse paths, we are ensuring the data reflects real-world spatiotemporal challenges. This not only mimics the dynamic nature of real-world scenarios but also enhances the application in localization and place recognition tasks. Additionally, the HeLiPR dataset overlaps with the MulRan dataset for long-term place recognition, presenting novel challenges in place recognition. With these features, the HeLiPR dataset is poised to become a valuable resource for improving place recognition and robotics applications, promoting advancements in the field.

\begin{acks}
This research was conducted with the support of the "National R\&D Project for Smart Construction Technology (24SMIP-A158708-05)" funded by the Korea Agency for Infrastructure Technology Advancement under the Ministry of Land, Infrastructure and Transport, and managed by the Korea Expressway Corporation.
\end{acks}

\bibliographystyle{SageH}
\bibliography{string-long, reference}

\begin{thebibliography}{31}
\providecommand{\natexlab}[1]{#1}
\providecommand{\url}[1]{\texttt{#1}}
\providecommand{\urlprefix}{URL }
\expandafter\ifx\csname urlstyle\endcsname\relax
  \providecommand{\doi}[1]{DOI:\discretionary{}{}{}#1}\else
  \providecommand{\doi}{DOI:\discretionary{}{}{}\begingroup
  \urlstyle{rm}\Url}\fi

\bibitem[{Agarwal et~al.(2020)Agarwal, Vora, Pandey, Williams, Kourous and
  McBride}]{agarwal2020ford}
Agarwal S, Vora A, Pandey G, Williams W, Kourous H and McBride J (2020) Ford
  multi-av seasonal dataset.
\newblock \emph{International Journal of Robotics Research} 39(12): 1367--1376.

\bibitem[{Arandjelovic et~al.(2016)Arandjelovic, Gronat, Torii, Pajdla and
  Sivic}]{arandjelovic2016netvlad}
Arandjelovic R, Gronat P, Torii A, Pajdla T and Sivic J (2016) Netvlad: Cnn
  architecture for weakly supervised place recognition.
\newblock In: \emph{Proceedings of the {IEEE} Conference on Computer Vision and
  Pattern Recognition}. pp. 5297--5307.

\bibitem[{Barnes et~al.(2020)Barnes, Gadd, Murcutt, Newman and
  Posner}]{RadarRobotCarDatasetICRA2020}
Barnes D, Gadd M, Murcutt P, Newman P and Posner I (2020) The oxford radar
  robotcar dataset: A radar extension to the oxford robotcar dataset.
\newblock In: \emph{Proceedings of the {IEEE} International Conference on
  Robotics and Automation}.

\bibitem[{Burnett et~al.(2023)Burnett, Yoon, Wu, Li, Zhang, Lu, Qian, Tseng,
  Lambert, Leung, Schoellig and Barfoot}]{Boreas}
Burnett K, Yoon DJ, Wu Y, Li AZ, Zhang H, Lu S, Qian J, Tseng WK, Lambert A,
  Leung KY, Schoellig AP and Barfoot TD (2023) Boreas: A multi-season
  autonomous driving dataset.
\newblock \emph{International Journal of Robotics Research} 42(1-2): 33--42.

\bibitem[{Chen et~al.(2022)Chen, Lopez, Agha-mohammadi and
  Mehta}]{chen2022direct}
Chen K, Lopez BT, Agha-mohammadi Aa and Mehta A (2022) Direct lidar odometry:
  Fast localization with dense point clouds.
\newblock \emph{{IEEE} Robotics and Automation Letters} 7(2): 2000--2007.

\bibitem[{Chen et~al.(2020)Chen, L\"abe, Milioto, R\"ohling, Vysotska, Haag,
  Behley and Stachniss}]{chen2020rss}
Chen X, L\"abe T, Milioto A, R\"ohling T, Vysotska O, Haag A, Behley J and
  Stachniss C (2020) {OverlapNet: Loop Closing for LiDAR-based SLAM}.
\newblock In: \emph{Proceedings of the Robotics: Science \& Systems
  Conference}.

\bibitem[{Chung et~al.(2023)Chung, Kim, Lee and Kim}]{Pohang}
Chung D, Kim J, Lee C and Kim J (2023) Pohang canal dataset: A multimodal
  maritime dataset for autonomous navigation in restricted waters.
\newblock \emph{International Journal of Robotics Research} 0(0):
  02783649231191145.

\bibitem[{Geiger et~al.(2012)Geiger, Lenz and Urtasun}]{Geiger2012CVPR}
Geiger A, Lenz P and Urtasun R (2012) Are we ready for autonomous driving? the
  kitti vision benchmark suite.
\newblock In: \emph{Proceedings of the {IEEE} Conference on Computer Vision and
  Pattern Recognition}.

\bibitem[{Helmberger et~al.(2021)Helmberger, Morin, Berner, Kumar, Wang, Yue,
  Cioffi and Scaramuzza}]{helmberger2022hilti}
Helmberger M, Morin K, Berner B, Kumar N, Wang D, Yue Y, Cioffi G and
  Scaramuzza D (2021) The hilti slam challenge dataset.

\bibitem[{Hsu et~al.(2021)Hsu, Kubo, Wen, Chen, Liu, Suzuki and
  Meguro}]{hsu2021urbannav}
Hsu L, Kubo N, Wen W, Chen W, Liu Z, Suzuki T and Meguro J (2021) Urbannav: An
  open-sourced multisensory dataset for benchmarking positioning algorithms
  designed for urban areas.
\newblock In: \emph{ION GNSS+}. pp. 226--256.

\bibitem[{Jeong et~al.(2019)Jeong, Cho, Shin, Roh and Kim}]{jjeong-2019-ijrr}
Jeong J, Cho Y, Shin YS, Roh H and Kim A (2019) Complex urban dataset with
  multi-level sensors from highly diverse urban environments.
\newblock \emph{International Journal of Robotics Research} 38(6): 642--657.

\bibitem[{Jung et~al.(2023)Jung, Jung and Kim}]{jung2023asynchronous}
Jung M, Jung S and Kim A (2023) Asynchronous multiple lidar-inertial odometry
  using point-wise inter-lidar uncertainty propagation.
\newblock \emph{{IEEE} Robotics and Automation Letters} .

\bibitem[{Kim et~al.(2021)Kim, Choi and Kim}]{kim2021scan}
Kim G, Choi S and Kim A (2021) Scan context++: Structural place recognition
  robust to rotation and lateral variations in urban environments.
\newblock \emph{{IEEE} Transactions on Robotics} 38(3): 1856--1874.

\bibitem[{Kim et~al.(2020)Kim, Park, Cho, Jeong and Kim}]{9197298}
Kim G, Park YS, Cho Y, Jeong J and Kim A (2020) Mulran: Multimodal range
  dataset for urban place recognition.
\newblock In: \emph{Proceedings of the {IEEE} International Conference on
  Robotics and Automation}. pp. 6246--6253.

\bibitem[{Knights et~al.(2023)Knights, Vidanapathirana, Ramezani, Sridharan,
  Fookes and Moghadam}]{wild}
Knights J, Vidanapathirana K, Ramezani M, Sridharan S, Fookes C and Moghadam P
  (2023) Wild-places: A large-scale dataset for lidar place recognition in
  unstructured natural environments.
\newblock In: \emph{Proceedings of the {IEEE} International Conference on
  Robotics and Automation}.

\bibitem[{Kong et~al.(2020)Kong, Yang, Zhai, Zhao, Zeng, Wang, Liu, Li and
  Wen}]{kong2020semantic}
Kong X, Yang X, Zhai G, Zhao X, Zeng X, Wang M, Liu Y, Li W and Wen F (2020)
  Semantic graph based place recognition for 3d point clouds.
\newblock In: \emph{Proceedings of the {IEEE}/{RSJ} International Conference on
  Intelligent Robots and Systems}. IEEE, pp. 8216--8223.

\bibitem[{Lee and Kim(2021)}]{9635907}
Lee AJ and Kim A (2021) Eventvlad: Visual place recognition with reconstructed
  edges from event cameras.
\newblock In: \emph{Proceedings of the {IEEE}/{RSJ} International Conference on
  Intelligent Robots and Systems}. pp. 2247--2252.

\bibitem[{Liu et~al.(2022)Liu, Yuan and Zhang}]{9779777}
Liu X, Yuan C and Zhang F (2022) Targetless extrinsic calibration of multiple
  small fov lidars and cameras using adaptive voxelization.
\newblock \emph{{IEEE} Transactions on Instrumentation and Measurement} 71:
  1--12.

\bibitem[{Luo et~al.(2021)Luo, Cao, Han, Shen and Li}]{9462410}
Luo L, Cao SY, Han B, Shen HL and Li J (2021) Bvmatch: Lidar-based place
  recognition using bird's-eye view images.
\newblock \emph{{IEEE} Robotics and Automation Letters} 6(3): 6076--6083.

\bibitem[{Mueggler et~al.(2018)Mueggler, Gallego, Rebecq and
  Scaramuzza}]{mueggler2018continuous}
Mueggler E, Gallego G, Rebecq H and Scaramuzza D (2018) Continuous-time
  visual-inertial odometry for event cameras.
\newblock \emph{{IEEE} Transactions on Robotics} 34(6): 1425--1440.

\bibitem[{Nguyen et~al.(2022)Nguyen, Yuan, Cao, Lyu, Nguyen and Xie}]{Viral}
Nguyen TM, Yuan S, Cao M, Lyu Y, Nguyen TH and Xie L (2022) Ntu viral: A
  visual-inertial-ranging-lidar dataset, from an aerial vehicle viewpoint.
\newblock \emph{International Journal of Robotics Research} 41(3): 270--280.

\bibitem[{Qingqing et~al.(2022)Qingqing, Xianjia, Queralta and
  Westerlund}]{qingqing2022multi}
Qingqing L, Xianjia Y, Queralta JP and Westerlund T (2022) Multi-modal lidar
  dataset for benchmarking general-purpose localization and mapping algorithms.
\newblock In: \emph{Proceedings of the {IEEE}/{RSJ} International Conference on
  Intelligent Robots and Systems}. pp. 3837--3844.

\bibitem[{Shan et~al.(2021)Shan, Englot, Duarte, Ratti and
  Rus}]{shan2021robust}
Shan T, Englot B, Duarte F, Ratti C and Rus D (2021) Robust place recognition
  using an imaging lidar.
\newblock In: \emph{Proceedings of the {IEEE} International Conference on
  Robotics and Automation}. pp. 5469--5475.

\bibitem[{Vidanapathirana et~al.(2021)Vidanapathirana, Moghadam, Harwood, Zhao,
  Sridharan and Fookes}]{vidanapathirana2021locus}
Vidanapathirana K, Moghadam P, Harwood B, Zhao M, Sridharan S and Fookes C
  (2021) Locus: Lidar-based place recognition using spatiotemporal higher-order
  pooling.
\newblock In: \emph{Proceedings of the {IEEE} International Conference on
  Robotics and Automation}. IEEE, pp. 5075--5081.

\bibitem[{Vidanapathirana et~al.(2022)Vidanapathirana, Ramezani, Moghadam,
  Sridharan and Fookes}]{vid2022logg3d}
Vidanapathirana K, Ramezani M, Moghadam P, Sridharan S and Fookes C (2022)
  Logg3d-net: Locally guided global descriptor learning for 3d place
  recognition.
\newblock In: \emph{Proceedings of the {IEEE} International Conference on
  Robotics and Automation}. pp. 2215--2221.

\bibitem[{{Wang} et~al.(2020){Wang}, {Wang} and {Xie}}]{wang2020intensity}
{Wang} H, {Wang} C and {Xie} L (2020) Intensity scan context: Coding intensity
  and geometry relations for loop closure detection.
\newblock In: \emph{Proceedings of the {IEEE} International Conference on
  Robotics and Automation}. pp. 2095--2101.

\bibitem[{Xu et~al.(2022{\natexlab{a}})Xu, Cai, He, Lin and Zhang}]{9697912}
Xu W, Cai Y, He D, Lin J and Zhang F (2022{\natexlab{a}}) Fast-lio2: Fast
  direct lidar-inertial odometry.
\newblock \emph{{IEEE} Transactions on Robotics} 38(4): 2053--2073.

\bibitem[{Xu et~al.(2022{\natexlab{b}})Xu, Lu, Wu, Lu, Zhu, Liao, Xiong and
  Wang}]{9981308}
Xu X, Lu S, Wu J, Lu H, Zhu Q, Liao Y, Xiong R and Wang Y (2022{\natexlab{b}})
  Ring++: Roto-translation invariant gram for global localization on a sparse
  scan map.
\newblock \emph{arXiv preprint arXiv:2210.05984} .

\bibitem[{Yuan et~al.(2024)Yuan, Lin, Liu, Wei, Hong and Zhang}]{10388464}
Yuan C, Lin J, Liu Z, Wei H, Hong X and Zhang F (2024) Btc: A binary and
  triangle combined descriptor for 3-d place recognition.
\newblock \emph{{IEEE} Transactions on Robotics} 40: 1580--1599.

\bibitem[{Yuan et~al.(2023)Yuan, Lin, Zou, Hong and Zhang}]{2023STD}
Yuan C, Lin J, Zou Z, Hong X and Zhang F (2023) Std: Stable triangle descriptor
  for 3d place recognition.
\newblock In: \emph{Proceedings of the {IEEE} International Conference on
  Robotics and Automation}.

\bibitem[{Zhang et~al.(2010)Zhang, Jin and Zhou}]{zhang2010understanding}
Zhang Y, Jin R and Zhou ZH (2010) Understanding bag-of-words model: a
  statistical framework.
\newblock \emph{International journal of machine learning and cybernetics} 1:
  43--52.

\end{thebibliography}

\end{document}